\documentclass{article} 
\usepackage[preprint]{colm2026_conference}

\usepackage{microtype}
\usepackage{hyperref}
\usepackage{url}
\usepackage{booktabs}
\usepackage{graphicx}
\usepackage{multicol}
\usepackage{enumitem}
\usepackage{xcolor}
\usepackage{tikz}
\usepackage{array}
\usetikzlibrary{arrows.meta,positioning,fit,backgrounds,calc,shadows.blur}
\usepackage{helvet}
\usepackage{siunitx}
\usepackage{multirow}
\usepackage[margin=1in]{geometry}
\usepackage[ruled,lined,linesnumbered,commentsnumbered]{algorithm2e}
\usepackage{amsmath,amssymb}
\usepackage{lmodern}
\usepackage[T1]{fontenc}
\usepackage{pgfplots}
\usepackage{tcolorbox}
\tcbuselibrary{breakable, skins}

\SetAlgoLined
\SetKwInOut{Input}{Input}
\SetKwInOut{Output}{Output}
\SetKwFunction{Convert}{ConvertToMarkdown}
\SetKwFunction{Clean}{CleanAndNormalise}
\SetKwFunction{Split}{SplitBySections}
\SetKwFunction{Index}{IndexByID}
\SetKwFunction{Hash}{SHA256}
\SetKwFunction{Generate}{LLMGenerate}
\SetKwFunction{Judge}{JudgeStage}
\SetKwFunction{Mean}{mean}
\SetKwProg{Fn}{Function}{:}{}
\DontPrintSemicolon
\SetArgSty{textnormal}

\usepackage{lineno}

\definecolor{darkblue}{rgb}{0, 0, 0.5}
\hypersetup{colorlinks=true, citecolor=darkblue, linkcolor=darkblue, urlcolor=darkblue}

\title{De~Jure: Iterative LLM Self-Refinement for Structured Extraction of Regulatory Rules}

\author{Keerat Guliani$^\dagger$, Deepkamal Gill$^\dagger$, David Landsman, Nima Eshraghi, Krishna Kumar \& Lovedeep Gondara \\
\small $^\dagger$Equal contribution 
\\
The Vanguard Group, Inc. \thanks{© 2026 The Vanguard Group, Inc. All rights reserved. \\This material is provided for informational purposes only and is not intended to be investment advice or a recommendation to take any particular investment action.}\\
{\small\texttt{\{keerat\_guliani, deepkamal\_gill,david\_landsman,nima\_eshraghi,krishna\_kumar,lovedeep\_gondara\}@vanguard.com}}
}
\definecolor{ink}{HTML}{2C3E50}
\definecolor{mute}{HTML}{7F8C9B}
\definecolor{cSrc}{HTML}{F5F6FA}   \definecolor{cSrcH}{HTML}{95A5A6}
\definecolor{cPre}{HTML}{D6EAF8}   \definecolor{cPreH}{HTML}{5DADE2}
\definecolor{cGen}{HTML}{D5F5E3}   \definecolor{cGenH}{HTML}{58D68D}
\definecolor{cJudge}{HTML}{FDEBD0} \definecolor{cJudgeH}{HTML}{F0B27A}
\definecolor{cRepair}{HTML}{FADBD8}\definecolor{cRepairH}{HTML}{EC7063}
\definecolor{cOut}{HTML}{E8DAEF}   \definecolor{cOutH}{HTML}{AF7AC5}
\definecolor{cPanel}{HTML}{FAFCFF}
\definecolor{cSchBg}{HTML}{F7F9FC} \definecolor{cSchH}{HTML}{5B6B7D}

\newlength{\hdrht}
\newlength{\cardht}

\tikzset{
  card/.style={
    rounded corners=2.5pt, draw=ink!20, line width=0.45pt,
    blur shadow={shadow blur steps=5, shadow xshift=0.3pt,
      shadow yshift=-0.5pt, shadow blur radius=1pt, shadow opacity=15},
    font=\fontsize{6}{7.2}\selectfont\sffamily, align=left,
    inner sep=0pt, outer sep=0pt,
    minimum height=\cardht          
  },
  hdr/.style={
    font=\fontsize{5.5}{6.5}\selectfont\sffamily\bfseries, text=white,
    inner sep=2.4pt, anchor=north west, minimum height=\hdrht
  },
  body/.style={
    anchor=north west, font=\fontsize{5}{6.2}\selectfont\sffamily,
    text=ink, align=left, inner sep=2.8pt
  },
  edge/.style={-{Stealth[length=2.2pt,width=1.7pt]},
    line width=0.55pt, draw=ink!65},
  retry/.style={-{Stealth[length=2pt,width=1.6pt]},
    line width=0.45pt, draw=cRepairH!80!black, densely dashed},
  eval/.style={-{Stealth[length=1.8pt,width=1.4pt]},
    line width=0.38pt, draw=mute, densely dotted},
  lbl/.style={font=\fontsize{4.2}{5}\selectfont\sffamily\itshape, text=mute},
  num/.style={circle, fill=ink, text=white,
    font=\fontsize{3.8}{4.5}\selectfont\sffamily\bfseries,
    inner sep=0.8pt, minimum size=3mm},
  mark/.style={circle, draw=ink!30, line width=0.35pt, fill=#1,
    minimum size=2.8mm, inner sep=0pt}
}

\begin{document}

\ifcolmsubmission
\linenumbers
\fi

\maketitle

\begin{abstract}
Regulatory documents encode legally binding obligations that LLM-based systems must respect. Yet converting dense, hierarchically structured legal text into machine-readable rules remains a costly, expert-intensive process. We present \textbf{De~Jure}, a fully automated, domain-agnostic pipeline for extracting structured regulatory rules from raw documents, requiring no human annotation, domain-specific prompting, or annotated gold data. De~Jure operates through four sequential stages: normalization of source documents into structured Markdown; 
LLM-driven semantic decomposition into structured rule units; multi-criteria LLM-as-a-judge evaluation across 19 dimensions spanning metadata, definitions, and rule semantics; and iterative repair of low-scoring extractions within a bounded regeneration budget, where upstream components are repaired before rule units are
evaluated, ensuring definitional context is maximally accurate before the most demanding decomposition stage. We evaluate De~Jure across four models on three regulatory corpora spanning finance, healthcare, and AI governance. On the finance domain, De~Jure yields consistent and monotonic increase in extraction quality, reaching peak performance within at most three judge-guided iterations. De~Jure generalizes effectively to the structurally distinct healthcare and the AI governance domains, maintaining similar high performance across all domains and both open- and closed-source models. In a downstream evaluation, using compliance question answering via RAG, responses grounded in De~Jure extracted rules are preferred by a judge LLM over prior work in 73.8\% of cases at single-rule retrieval depth, rising to 84.0\% under broader retrieval, confirming that extraction fidelity translates directly into downstream utility. These results demonstrate that explicit, interpretable evaluation criteria can substitute for human annotation in highly complex regulatory domains, offering a scalable and auditable path toward regulation-grounded LLM alignment.

\end{abstract}

\section{Introduction}
\label{sec:intro}
The alignment of large language models (LLMs) with human values has become one of the central challenges in modern AI research~\citep{ouyang2022instructgpt,bai2022hhrlhf}.
While most alignment work focuses on making models helpful and
harmless~\citep{bai2022hhrlhf,christiano2017rlhf}, the rapid deployment of LLMs in
high-stakes domains including finance~\citep{wu2023bloomberggpt,yang2023fingpt},
healthcare~\citep{singhal2023med}, and law~\citep{chalkidis2020legalbert,nay2023law}
demands a broader conception of alignment: one grounded not only in subjective human
preferences but in explicit, codified regulatory obligations. Regulatory documents such
as healthcare privacy rules (\textsc{HIPAA}~\citep{hipaa}), financial conduct standards (\textsc{SEC Advisers Act}~\citep{sec}), and
responsible AI frameworks (\textsc{EU AI Act}~\citep{euaiact}) encode legally binding requirements that
LLM-based systems must respect. Yet transforming these dense, hierarchically structured
texts into machine-readable rules remains a largely manual, expert-intensive process,
creating a critical bottleneck for compliance-aware AI deployment at scale.

Constitutional AI (CAI)~\citep{bai2022constitutional} and its
extensions~\citep{lee2023rlaif,sun2023salmon} have demonstrated that LLMs can generate and evaluate \emph{alignment principles} for general helpfulness and harmlessness from plain-language policy text, reducing reliance on costly human preference annotation. However, in high-stakes domains such as finance, healthcare, and law, alignment is governed by explicit regulatory obligations rather than general-purpose safety principles, which are often structurally complex and exception-laden. Existing approaches either depend on hand-curated seed principles requiring significant domain expertise~\citep{bai2022constitutional}, or produce coarse-grained rules ill-suited for the structural complexity of legal and regulatory corpora~\citep{sun2023salmon}.
More targeted work on LLM-driven legal rule extraction has shown promise. Defeasible deontic logic formulae have been extracted from telecommunications  regulations~\citep{governatori2022norm}, and multi-stage pipelines have been applied to regulatory text~\citep{sleimi2018automated}. The work most closely related to ours \citep{datla2025executable}, extracts governance principles from \textsc{HIPAA} and the \textsc{EU AI Act} via few-shot prompting with judge-and-repair steps benchmarked against a human-annotated gold set. However, it requires costly expert annotation, captures a relatively flat rule representation without decomposing the finer-grained semantic structure of regulatory obligations, and treats repair as a single, undifferentiated correction step without exploiting the hierarchical dependencies between section metadata, term definitions, and rule units. 

\begin{figure*}[t]
  \centering

  \includegraphics[width=\linewidth]{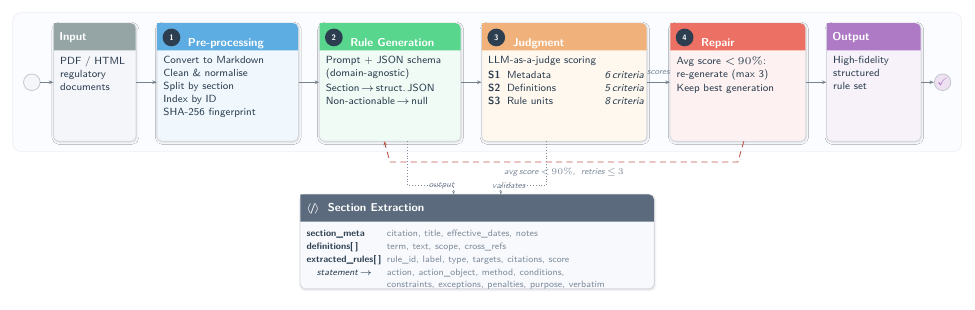}

  \caption{De Jure pipeline overview.
    Input documents are pre-processed into structured Markdown (Stage~1),
    parsed into JSON rule units via a domain-agnostic LLM prompt (Stage~2),
    scored by an LLM judge across 19 criteria in three stages (Stage~3),
    and iteratively repaired until the per-stage average score reaches
    90\% or the retry budget (max~3) is exhausted (Stage~4).}\label{fig:main_method}
\end{figure*}

We propose \textbf{De~Jure} (\textbf{D}ocument \textbf{E}xtraction with
\textbf{Ju}dge-\textbf{R}efined \textbf{E}valuation) (Figure~\ref{fig:main_method}),
a fully automated, domain-agnostic pipeline that transforms raw regulatory documents
into structured, machine-readable rule sets with no human annotation and no
domain-specific prompts. De~Jure pre-processes input documents into normalized
Markdown, prompts an LLM to extract typed rule units conforming to a schema that
decomposes each rule into a rich set of semantic fields capturing actions, conditions,
constraints, exceptions, penalties, and verbatim source spans, and applies a
multi-criteria LLM-as-a-judge~\citep{zheng2023judging} to score each extraction
across 19 dimensions~(summarized in Appendix~\ref{sec:appx_judge_criteria}). A key design principle is \emph{hierarchical decoupling}, where the components that rules depend on are verified and repaired before rule units are evaluated, ensuring that rule-level repair always operates on reliable context. Stages that fall below a quality threshold of 90\% are repaired through targeted regeneration, and the highest-scoring output is retained within a configurable user-fined retry budget (defaulting to three attempts). Our ablation studies confirm that this budget is both necessary and sufficient.
Our main contributions are as follows:

\begin{itemize}

  \item \textbf{De~Jure pipeline and extraction schema.} A four-stage, domain-agnostic
  pipeline for end-to-end regulatory rule extraction requiring no annotated data,
  domain-specific prompts, or logical formalisms, anchored by a structured schema
  that decomposes each rule unit into a rich set of semantic fields generalizable across
  regulatory corpora without modification (Section~\ref{sec:main_method}).

\item \textbf{Multi-stage LLM judge with hierarchical repair.} A 19-criterion evaluation framework with three judges applied in hierarchical dependency order, so that each stage repairs on previously-verified context before the next begins. A targeted regeneration mechanism retains the highest-scoring output within a bounded compute budget (Section~\ref{jr}).

\item \textbf{Cross-domain generalization.} With no changes to prompts, schema,
or model configuration, De~Jure generalizes across three structurally distinct
regulatory domains: \textsc{HIPAA} (healthcare), the \textsc{SEC Advisers Act}
(finance), and the \textsc{EU AI Act} (AI governance), achieving total average
scores above 94\% (4.70/5.00) across all domain and model combinations. This demonstrates that the pipeline transfers to new regulatory domains without any re-engineering
(Section~\ref{sec:exp2}).

 \item \textbf{Downstream and ablation evaluation.} A RAG-based question answering comparison against \citet{datla2025executable} on \textsc{HIPAA} shows that De~Jure-grounded responses
are preferred in 73.8\% of cases at single-rule retrieval, rising to 84.0\% at
ten-rule retrieval, confirming that extraction quality gains translate directly to
downstream utility. Targeted ablation studies validate the contribution of each
key design choice to overall pipeline quality
(Section~\ref{sec:downstream-eval}, Appendix~\ref{sec:app_ablation}).

\end{itemize}

The remainder of this paper is organized as follows. Section~\ref{sec:related} reviews related work. Section~\ref{sec:main_method} describes the De~Jure pipeline, covering
preprocessing, extraction schema, judge design, and iterative repair. Section~\ref{sec:experiments} presents our evaluation, including extraction quality, cross-domain generalization, and downstream RAG assessment. Finally, the concluding remarks are given in Section~\ref{sec:conclusion}. 
Supplementary material is provided in the appendix, including ablation studies 
(Appendix~\ref{sec:app_ablation}), implementation and algorithmic details (Appendices~\ref{sec:appx_judge_criteria} and~\ref{appx:alg_diag}), a qualitative 
end-to-end pipeline case study (Appendix~\ref{appx:example}), extended experimental 
results (Appendices~\ref{appx:extraction_quality_results} and~\ref{appx:gen_results}), 
and complete prompts and extraction schemas (Appendices~\ref{appx:prompts} and~\ref{appx:schema}).


\section{Related Work}
\label{sec:related}

De~Jure sits at the intersection of regulatory NLP, structured information extraction, and LLM-based self-refinement. We review the work most relevant to these three areas and clarify where our approach diverges from prior methods.

\subsection{Rule Extraction from Regulatory and Policy Text}
\label{sec:related_extraction}
Converting dense regulatory text into structured, machine-readable representations is a
long-standing problem in legal informatics, healthcare, and finance. Early work applied
syntactic parsing and logic-based formalisms to identify normative statements and support
automated compliance checking~\citep{dragoni2016combining,wyner2011rule,governatori2024asp},
while information extraction approaches have been applied to financial
regulations~\citep{lam2021extracting} and clinical guidelines~\citep{weng2010extracting}.
More recent work leverages LLMs for rule extraction across domains, including traffic
regulations~\citep{zin2025towards}, clinical decision support~\citep{he2024generative,tang2026policy},
and legal contract understanding~\citep{koreeda2021contractnli,hendrycks2021cuad}. Despite this
progress, existing approaches either require labeled data, operate within a single domain,
or produce extractions too coarse to capture the conditional and exception-laden structure
of regulatory obligations. De~Jure addresses all three limitations simultaneously.

\subsection{Deontic Logic and Defeasible Rule Formalisms}
\label{sec:related_deontic}

A dominant paradigm for making extracted rules machine-actionable is \emph{deontic
logic}~\citep{vonwright1951deontic}, extended by \emph{defeasible deontic logic} (DDL)
to support reasoning under exceptions and conflicts~\citep{governatori2023deontic}.
At scale, DDL extraction has been applied to telecommunications regulations using
carefully designed prompts and fine-tuning~\citep{horner2025toward}. While powerful,
DDL-based extraction is inherently domain-specific: the target formalism must be defined
in advance and does not transfer to new regulatory domains without substantial
re-engineering.


\subsection{LLM pipelines for structured principle extraction and iterative refinement}
\label{sec:related_pipelines}
CLAUDETTE~\citep{lippi2019claudette} and OPP-115~\citep{wilson2016creation} established
that structured, criterion-level analysis of policy text is both feasible and practically
valuable. The most closely related work, of ~\citet{datla2025executable} extracts governance
principles from texts including \textsc{HIPAA} and the \textsc{EU AI Act} via chunking, clause mining, and
structured LLM extraction with judge-and-repair steps evaluated against a human-annotated
gold set. On the refinement side, iterative self-refinement~\citep{madaan2023selfrefine},
verbal reinforcement~\citep{shinn2023reflexion}, and Constitutional
AI~\citep{bai2022constitutional} have demonstrated that LLM-generated critiques can
systematically improve output quality without annotated data, and LLM-as-a-judge
evaluation with decomposed criteria has been shown to correlate more closely with human
judgment than holistic scoring~\citep{zheng2023judging,liu2023geval}.

\paragraph{Positioning De~Jure.} 
Prior approaches address complementary aspects of the problem but differ from
De~Jure in three key respects. DDL-based methods~\citep{horner2025toward}
require a pre-specified logical formalism and do not transfer across domains. The method in 
\citet{datla2025executable} depends on costly human annotation and applies repair
as a single flat step, without accounting for structural dependencies in the
extraction. General-purpose refinement methods~\citep{madaan2023selfrefine,shinn2023reflexion}
provide no structured quality criteria and are not designed for regulatory text.
De~Jure addresses all three limitations. It requires no formal specification and
no human annotation. Furthermore, repair steps are applied hierarchically, so that
upstream components are corrected before rule units are evaluated. Finally,
criterion-driven, field-level judgment at each stage enables scalable quality
control across heterogeneous regulatory corpora with minimal domain expertise.

\section{De Jure} \label{sec:main_method}

We present \textbf{De~Jure} (\textbf{D}ocument \textbf{E}xtraction with
\textbf{Ju}dge-\textbf{R}efined \textbf{E}valuation), an automated,
domain-agnostic pipeline for transforming raw regulatory documents into structured,
machine-readable rule sets (Figure~\ref{fig:main_method}). Unlike prior works that
rely on human-annotated gold sets~\citep{sleimi2018automated} or domain-specific
logical formalisms~\citep{governatori2022norm}, De~Jure requires no labelled data and
no domain expertise beyond the source document itself. A full procedural specification
of the pipeline is provided in Appendix~\ref{appx:alg_diag}
(Algorithm~\ref{alg:pipeline}), and a concrete end-to-end example of De~Jure in
action is given in Appendix~\ref{appx:example} (Figure~\ref{fig:example}).

The pipeline consists of four stages: (1)~document \emph{pre-processing} into
normalized, section-segmented Markdown; (2)~LLM-driven \emph{rule generation} into a
typed JSON schema capturing a rich set of semantic fields per rule unit; (3)~\emph{multi-criteria
judgment} by an LLM judge across 19 dimensions (summarized in Appendix~\ref{sec:appx_judge_criteria}) organized in three sequential validation
stages; and (4)~\emph{selective repair by regeneration}, where any stage scoring below a
quality threshold is retried and the highest-scoring output is retained. Two core principles drive the De~Jure pipeline design. First, \emph{decoupling of extraction from verification}: by separating what is generated from how it is evaluated, the same judgment-repair loop
applies uniformly across document types, regulatory regimes, and LLMs without
modifying the core pipeline. Second, \emph{hierarchical repair ordering}: the three judgment stages are applied in dependency order, so that upstream components are verified and repaired before rule units are evaluated, ensuring that rule-level repair always operates
on reliable context.

\subsection{Pre-processing} \label{pp}

The pre-processing stage (Figure~\ref{fig:main_method}, Stage~1) converts
heterogeneous regulatory source formats into structured section--content pairs
suitable for downstream LLM processing. Each input document is converted to
Markdown using \texttt{Docling}~\citep{docling2024}, which preserves section
boundaries, list structure, and table formatting, then cleaned to remove
formatting artifacts and extraneous whitespace. The clean text is segmented
by splitting on regulatory section delimiters (e.g., ``\S'', ``Article'',
``Rule''), yielding an ordered set $\mathcal{S}$ of section--content pairs,
each indexed by its regulatory identifier, assigned a SHA-256 fingerprint for
traceability, and stored alongside standardized metadata fields (title, version,
effective dates). This deterministic representation ensures every downstream
extraction traces back to an exact source span, a hard requirement for
regulatory auditability.



\subsection{Rule Generation} \label{rg}

Each section $s \in \mathcal{S}$ is independently submitted to the rule generation
stage (Figure~\ref{fig:main_method}, Stage~2), where an LLM is prompted to produce
a structured JSON extraction conforming to our \texttt{Section Extraction} schema
(Figure~\ref{fig:main_method}). The schema decomposes each section into three typed
components: (i)~\emph{section metadata} (citation, title, effective dates, notes);
(ii)~\emph{definitions} (term, text, scope, cross-references); and (iii)~\emph{rule
units}, each carrying an identifier, rule type, summary label, citation, and a
nine-field statement decomposition: action, action object, method, conditions,
constraints, exceptions, penalties, purpose, and verbatim source span. This
fine-grained decomposition enables downstream systems: constitutional AI
frameworks~\citep{bai2022constitutional} or compliance verification engines to
query and enforce individual semantic components rather than treating rules as
monolithic strings. The generation prompt is schema-driven with no domain-specific
examples or seed rules, and returns \texttt{null} for non-actionable sections,
suppressing non-normative passages such as preambles and cross-reference tables to
prevent rule set inflation. Full prompt and schema are provided in Appendix~\ref{appx:prompts} and Appendix~\ref{appx:schema}, respectively.

\subsection{Multi-Criteria Judgment}
\label{jr}

Since De~Jure operates in an \emph{unsupervised} setting with no reference annotations,
quality control cannot rely on surface-form matching against a gold standard. Instead,
we adopt an LLM-as-a-judge framework~\citep{zheng2023judging, liu-etal-2023-g, zhong-etal-2022-towards} organized into
\emph{three sequential validation stages} mirroring the schema hierarchy: section metadata is
judged and repaired first, followed by definitions, then rule units. This ordering
ensures each stage receives the benefit of all prior corrections. By the time rule units are evaluated, both the section metadata and the definitional vocabulary have already been refined, maximizing the contextual accuracy available to the rule repair stage and increasing the likelihood of producing high-fidelity rule decompositions.


\textbf{Stage 1: Metadata validation (6 criteria).}
We validate section-level metadata including headings, effective dates, citation
strings, and optional contextual fields across six criteria: completeness, fidelity to source, non-hallucination, title quality, citation and date precision, and optional field population. Accurate metadata is foundational: it anchors every downstream extraction to a traceable regulatory provision, and errors such as incorrect effective dates or misattributed citations propagate silently into all derived rule units.

\textbf{Stage 2: Definition validation (5 criteria).}
We validate all extracted definitions, covering both affirmative
(``X means Y'') and exclusionary (``X does not include Y'') forms. Where a
definitional span also encodes an operative rule, it must appear in both the
definitions and rule-unit components and is evaluated independently in each. Criteria include completeness, source fidelity, non-hallucination, precision and formatting, and term quality. Non-hallucination is particularly critical here: a fabricated or paraphrased definition silently corrupts the interpretation of every rule referencing that term, making this the highest-leverage point for factual grounding in the pipeline.

\textbf{Stage 3: Rule-unit validation (8 criteria).}
We validate each rule unit at the field level across criteria spanning the full
statement decomposition: completeness, label conciseness, rule-type
classification accuracy, fidelity to source, neutrality, target consistency,
actionability, and non-hallucination. Several criteria warrant brief motivation.
\emph{Rule-type accuracy} is essential because misclassifying an obligation as a
permission silently inverts compliance semantics. \emph{Neutrality} prevents the model
from injecting interpretive framing that would bias downstream enforcement.
\emph{Actionability} ensures each rule is expressed in a form that a compliance system can
operationalize. \emph{Label conciseness} matters because summary labels serve as
retrieval keys in downstream question answering RAG systems, where verbose or ambiguous labels directly
degrade retrieval precision. This is the most demanding stage: a single source section
may yield multiple rule units, each required to satisfy all criteria independently.

Each criterion is scored on a $0$--$5$ scale by the judge LLM, which also produces a
natural-language justification for each score. The full set of criteria is summarized in
Table~\ref{tab:judgement-criteria}, Appendix~\ref{sec:appx_judge_criteria}. 


\subsection{Selective Repair by Regeneration}\label{sec:repair}
Low judgment scores indicate specific, localizable defects such as a misclassified rule
type, an incomplete label, a hallucinated condition rather than wholesale
generation failure. De~Jure exploits this through a \emph{selective repair} mechanism:
each stage is repaired independently if and only if its average score falls below
$\theta = 0.90$. The LLM is re-prompted with the original section text, the current
extraction, per-criterion scores, and the judge's natural-language critiques, and asked
to correct only the deficient fields. This repeats for up to $r{=}3$ attempts per
stage, retaining the best-scoring output across all attempts, guaranteeing
monotonically non-decreasing quality.

This design has three practical consequences. \emph{First}, repair cost is bounded, with at most $r$ additional LLM calls per stage per section. \emph{Second}, structured scores and critiques make the repair signal substantially richer than a generic re-prompt, empirically producing targeted field-level corrections rather than wholesale rewrites. \emph{Third}, retaining the best rather than the final generation provides a soft safety net: when re-generation degrades quality, the pipeline falls back gracefully to the best previously seen output.




\section{Experiments and Results}
\label{sec:experiments}
We evaluate De~Jure across four backbone extraction and repair models with a separate fixed judge model, on three regulatory corpora,
assessing extraction quality, cross-domain generalization, downstream utility,
and core design decisions through ablation studies.

\paragraph{Pipeline Configuration.}
Extraction and judgment are performed by strictly separate models. Extraction
is performed by the backbone model under evaluation while judgment is performed by a fixed model,
Compass~\citep{compass}, designed for structured, criterion-aligned
evaluation of long-form outputs. Three judges operate sequentially, each
instantiated from Compass with stage-specific criteria targeting the failure
modes of its semantic layer: ~\textbf{Judge~1} evaluates section metadata
quality (6 criteria), \textbf{Judge~2} assesses extracted definitions
(5 criteria), and \textbf{Judge~3} evaluates the correctness and completeness of extracted rules (8 criteria). Each produces per-criterion scores and
natural-language critiques. If the average normalized score falls below $\theta = 90\%$, the
backbone model is re-prompted with the critique, the prior extraction, and
the original input, for at most $r{=}3$ attempts per judge, retaining the
highest-scoring output across all attempts.


\paragraph{Inference Settings.}
All backbone models use identical hyperparameters: temperature $\tau = 0.1$,
maximum output tokens $T_{\max} = 4096$, and nucleus sampling $p = 0.95$,
encouraging deterministic, faithful extraction.



\subsection{Datasets}\label{sec:datasets}
We evaluate on three regulatory corpora spanning distinct high-stakes domains
(Table~\ref{tab:datasets}), selected to probe generalization across diverse legal
styles and rule structures rather than in-domain memorization.


\paragraph{SEC Investment Advisers Act.}
The \textit{Investment Advisers Act}~\citep{sec} governs investment adviser conduct in
U.S.\ financial markets. Its densely nested conditionals and extensive
cross-references make it a demanding primary benchmark.

\paragraph{EU Artificial Intelligence Act.}
The \textit{Regulation (EU) 2024/1689}~\citep{euaiact} is the world's first
comprehensive AI governance framework. Its mix of binding technical obligations,
broad governance principles, and non-binding recitals presents a structurally
distinct extraction challenge.

\paragraph{HIPAA Privacy Rule.}
The \textit{HIPAA Privacy Rule}~\citep{hipaa} governs the use and disclosure of
protected health information in the United States. Its exception-laden permission
structures, where broad prohibitions are qualified by narrow condition-dependent
exemptions, test a pipeline's ability to faithfully decompose rule scope.

\begin{table}[h]
\centering
\resizebox{\textwidth}{!}{%
\begin{tabular}{llllc}
\toprule
\textbf{Dataset} & \textbf{Domain} & \textbf{Jurisdiction}
    & \textbf{Sections} & \textbf{Rule Density} \\
\midrule
SEC Advisers Act & Financial securities
    & United States  & $\sim$50  & High \\
EU AI Act    & AI governance
    & European Union & 113       & Medium--High \\
HIPAA Privacy Rule & Healthcare privacy
    & United States  & $\sim$30  & High \\
\bottomrule
\end{tabular}%
}
\caption{Summary of evaluation datasets.}
\label{tab:datasets}
\end{table}

The three corpora span a natural complexity gradient: \textsc{SEC} provisions are 
rigidly alphanumerically indexed with well-delimited boundaries; \textsc{HIPAA} 
organizes mandates around exception-laden permission hierarchies with less consistent 
demarcation; and the \textsc{EU AI Act} employs discursive, principle-based prose that 
interleaves binding obligations with non-binding recitals. Ordered from most to least 
structurally regular, they constitute an \emph{a priori} difficulty ranking for extraction.


\subsection{Experiment 1: Extraction Quality}\label{sec:exp1}

\paragraph{Experimental Setup.} 
We evaluate on the SEC Investment Advisers Act corpus using four backbone models
spanning open- and closed-source families: \textbf{Llama-3.1-8B-Instruct}~\citep{llama3}
and \textbf{Qwen3-VL-8B-Instruct}~\citep{qwen3} as cost-efficient, controllable
open-source options, and \textbf{Claude-3.5-Sonnet}~\citep{claude35} and
\textbf{GPT-5-mini}~\citep{gpt5mini} as high-capability closed-source systems.
This selection enables a controlled comparison across model families and access
paradigms under identical pipeline conditions.


\begin{table}[h]
\centering
\setlength{\tabcolsep}{8pt}
\begin{tabular}{ll S[table-format=1.2] S[table-format=1.2] S[table-format=1.2]}
\toprule
\textbf{Model}
    & {\textbf{J1 (Metadata)}}
    & {\textbf{J2 (Definitions)}}
    & {\textbf{J3 (Per-Rule)}}
    & {\textbf{Overall}} \\
\midrule
Llama-3.1-8B
    & 4.93
    & 4.75
    & 4.53
    & 4.74 \\
Qwen3-VL-8B
    & 4.93
    & \textbf{4.87}
    & 4.64
    & 4.81 \\
Claude-3.5-Sonnet
    & \textbf{4.99}
    & 4.83
    & 4.67
    & 4.83 \\
GPT-5-mini
    & \textbf{4.99}
    & 4.83
    & \textbf{4.75}
    & \textbf{4.85} \\
\bottomrule
\end{tabular}
\caption{Averaged judge (J) scores on the SEC Investment Advisers Act corpus
(scale 1--5; higher is better). Per-criterion breakdowns for all three judges
are provided in Appendix~\ref{appx:extraction_quality_results}
(Tables~\ref{tab:extraction_step1_criteria}--\ref{tab:extraction_step3_criteria}).}

\label{tab:sec_results}
\end{table}

\paragraph{Results.} 
Table~\ref{tab:sec_results} reports the averaged score across all judge criteria, for each model and judge stage.
Two observations stand out. First, performance degrades monotonically from metadata
($\approx$4.96) to definitions ($\approx$4.82) to per-rule quality ($\approx$4.65),
directly reflecting increasing task complexity: section-level metadata is structurally
salient and consistently recoverable, whereas decomposing fine-grained rule components
such as conditions, exceptions, and penalties demands substantially deeper semantic
parsing. Second, model divergence
is most pronounced at Judge 3 (J3), where GPT-5-mini leads (Avg~$= 4.75$) over Llama-3.1-8B
(4.53), driven by superior Accuracy (4.95) and Fidelity to Source (4.74) on individual
rule decomposition (Appendix~\ref{appx:extraction_quality_results}). Notably, Qwen3-VL-8B achieves the highest J2 average (4.87),
outperforming both closed-source models on definition extraction. The narrow overall
gap between the best open-source model (Qwen3-VL-8B, 4.81) and the best closed-source
model (GPT-5-mini, 4.85) suggests that capable open-source models, when paired with
structured extraction and iterative refinement, can approach proprietary model
performance, a practically significant finding for privacy-sensitive regulatory
deployments where reliance on external APIs is undesirable.
Further, the per-criterion breakdowns in Appendix~\ref{appx:extraction_quality_results}
reveal that Non-Hallucination is uniformly perfect (5.00) across all models and all
three judges, confirming that schema-constrained extraction eliminates factual
fabrication as a failure mode regardless of model family.


\subsection{Experiment 2: Generalization Across Regulatory Domains}\label{sec:exp2}
\paragraph{Experimental Setup.}
Does De~Jure generalize, \emph{without any modification}, across structurally diverse regulatory domains? We evaluate on all three corpora from Section~\ref{sec:datasets},
using \textbf{GPT-5-mini} (best closed-source) and \textbf{Qwen3-VL-8B-Instruct}
(best open-source) from Experiment~1, with all settings unchanged.




\begin{table}[h]
\centering
\setlength{\tabcolsep}{8pt}
\begin{tabular}{ll S[table-format=1.2] S[table-format=1.2] S[table-format=1.2] S[table-format=1.2]}
\toprule
\textbf{Dataset} & \textbf{Model}
    & {\textbf{J1 (Metadata)}}
    & {\textbf{J2 (Definitions)}}
    & {\textbf{J3 (Per-Rule)}}
    & {\textbf{Overall}} \\
\midrule
\multirow{2}{*}{SEC}
  & GPT-5-mini   & \textbf{4.99} & 4.83          & \textbf{4.75} & \textbf{4.85} \\
  & Qwen3-VL-8B  & 4.93          & \textbf{4.87} & 4.64          & 4.82          \\
\cmidrule(lr){1-6}
\multirow{2}{*}{HIPAA}
  & GPT-5-mini   & \textbf{4.93} & 4.61          & \textbf{4.75} & \textbf{4.76} \\
  & Qwen3-VL-8B  & 4.90          & \textbf{4.71} & 4.66          & 4.76          \\
\cmidrule(lr){1-6}
\multirow{2}{*}{EU AI Act}
  & GPT-5-mini   & \textbf{4.72} & 4.69          & \textbf{4.71} & 4.71          \\
  & Qwen3-VL-8B  & 4.65          & \textbf{4.82} & 4.65          & \textbf{4.71} \\
\bottomrule
\end{tabular}
\caption{
Generalization across three regulatory corpora (scale 1--5; higher is
better). De~Jure is applied without domain-specific modification. Per-criterion
breakdown details are provided in
Appx.~\ref{appx:gen_results} (Tables~\ref{tab:gen_step1_criteria}--\ref{tab:gen_step3_criteria}).}
\label{tab:generalization}
\end{table}

\paragraph{Results.}
Table~\ref{tab:generalization} shows a consistent result: overall scores remain above
4.70 across every domain and model combination, the range spanning only 0.14 points in total. 
Sustaining near-ceiling performance across three structurally distinct regulatory domains, with no domain-specific adaptation, presents strong evidence of broad domain generalizability.
The monotonic decline from SEC ($\approx$4.84)
to HIPAA ($\approx$4.76) to EU AI Act ($\approx$4.71) tracks an intuitive ordering of
structural regularity: SEC text is rigidly indexed, HIPAA is more loosely organized,
and the EU AI Act is the most discursive. 
This monotonic decline mirrors the \emph{a priori} structural complexity ordering 
established in Section~\ref{sec:datasets}, which was derived independently of any 
experimental result. A permissive judge would yield uniformly near-ceiling scores 
across all corpora; the systematic score variation is consistent only with a judge 
that is sensitive to intrinsic differences in extraction difficulty.
This finding is further supported by the qualitative evidence in Appendix~\ref{appx:example}, where the judge assigns a failing normalized average score of 0.55 to a semantically deficient extraction, with targeted per-criterion feedback that directly identifies the defective fields. Upon repair, the same judge awards a passing normalized average score of 0.90 to the corrected output, while leaving scores unchanged on fields that were already correct. This behavior confirms that the judge is discriminative rather than permissive. It penalizes specific deficiencies, recognizes genuine improvement, and does not reward outputs indiscriminately.

Two patterns at the averaged and at the per-criterion level (Table~\ref{tab:generalization} and Appendix~\ref{appx:gen_results})
merit attention. First, J2 is the most variable judge and the primary driver of
the cross-domain gap: GPT-5-mini's J2 drops from 4.83 (SEC) to 4.61 (HIPAA),
consistent with the intuition that linking rules to definitional context is harder
when section boundaries are loosely organized rather than rigidly indexed.
 Model-level variance is most pronounced at J2, with scores diverging by up to 0.21 points across domains, suggesting that definitional enrichment is sensitive to corpus structure and model characteristics.
Second, J3 remains the hardest stage across all three corpora,
consistent with Experiment~1, confirming that precise recovery of conditional
triggers, quantitative thresholds, and nested exception structures is a universal
bottleneck independent of jurisdiction or legal style, and identifying fine-grained
rule decomposition as the primary target for future work.

\subsection{Experiment 3: Downstream Evaluation, Compliance QA via RAG}
\label{sec:downstream-eval}
Intrinsic extraction quality does not fully capture practical utility. We therefore
evaluate extracted rules as the knowledge base of a RAG system tasked with answering
HIPAA compliance questions, providing a direct task-grounded comparison against
\citet{datla2025executable}.


\paragraph{Experimental Setup.}
We restrict evaluation to HIPAA sections covered by the publicly released extractions
of~\citet{datla2025executable},\footnote{\scriptsize\url{https://github.com/gautamvarmadatla/Policy-Tests-P2T-for-operationalizing-AI-governance/blob/90059a7d2b59d705a80d212b3a1a8adfb30ef58b/Annotator\%20Data/Annotator\%20Docs/out/HIPAA/HIPAA_removed.extracted.jsonl}}
ensuring any performance gap is attributable solely to rule quality rather than
coverage. Our rules are extracted using GPT-5-mini. We prompt Qwen3-VL-8b-Instruct to generate 100 evaluation questions from the same HIPAA sections at temperature $\tau{=}0.8$ (Further details in Appendices~\ref{appx:question_gen_prompt} and \ref{appx:sample_questions}), yielding a lexically and semantically diverse set in which every question has a verifiable, source-grounded answer. Two independent vector databases are constructed, one per rule set, both encoded with \texttt{all-mpnet-base-v2}~\citep{reimers-2019-sentence-bert}. Answers are generated using Claude~3.5~Sonnet under retrieval depths
$k \in \{1, 5, 10\}$, spanning precise single-rule to broad multi-rule regimes. A
pairwise LLM judge ($\tau{=}0.1$) evaluates each answer pair across six criteria:
Completeness, Factual Grounding, Handling Ambiguity, Practical Actionability,
Regulatory Precision, and Overall Preference (Appx.~\ref{appx:ds_judge_criteria} provides further details). To eliminate positional bias, each pair
is judged twice with swapped ordering and win rates are averaged.

\begin{table}[h]
\centering
\small
\setlength{\tabcolsep}{9pt}
\renewcommand{\arraystretch}{1.15}
\begin{tabular}{lccc}
\toprule
\textbf{Criterion} & $k{=}1$ & $k{=}5$ & $k{=}10$ \\
\midrule
Completeness            & 78.00 & 80.50 & 83.50 \\
Factual Grounding       & 80.50 & 76.00 & 85.50 \\
Handling Ambiguity      & 53.50 & 66.50 & 84.00 \\
Practical Actionability & 78.00 & 80.50 & 84.00 \\
Regulatory Precision    & 74.50 & 80.50 & 83.50 \\
Overall Preference      & 78.00 & 80.50 & 83.50 \\
\midrule
\textbf{Aggregated}     & \textbf{73.75} & \textbf{77.42} & \textbf{84.00} \\
\bottomrule
\end{tabular}
\caption{Pairwise win rates (\%) of De~Jure against~\citet{datla2025executable} on the
HIPAA downstream QA task across three retrieval depths.}
\label{tab:downstream_eval}
\end{table}

\paragraph{Results.}
Table~\ref{tab:downstream_eval} shows De~Jure outperforming \citet{datla2025executable}
across every criterion and retrieval depth, with a widening margin as $k$ grows. At
$k{=}1$, where aggregation effects are absent, the 73.75\% win rate, 23.75 points
above parity, reflects a fundamental representational advantage: our rules preserve
conditional logic, scope qualifiers, and exception clauses that the baseline's flatter
representations omit, with factual grounding already pronounced at 80.50\%. The most
diagnostic trajectory is handling ambiguity, which begins near parity at $k{=}1$
(53.50\%) and surges to 84.00\% at $k{=}10$ (+30.50 pts). Ambiguous queries are
inherently multi-provision and cannot be resolved by any single rule; that our advantage
materializes precisely as $k$ grows confirms that our structured decomposition produces
rules that integrate compositionally across provisions, while the baseline exhibits
diminishing returns consistent with redundant, less differentiated extractions. The
monotonically increasing aggregate win rate from 73.75\% to 77.42\% to 84.00\% further
confirms that De~Jure's rules are mutually complementary, each additional retrieved
rule contributes distinct, non-redundant context, and that extraction fidelity
translates directly into downstream utility.

\vspace{-0.3cm}
\section{Ablation Studies}\label{sec:ablation_summary}\vspace{-0.3cm}
We conduct four targeted ablations on core design decisions (full results in Appendix \ref{sec:app_ablation}). First, extraction quality improves monotonically with acceptance threshold $\theta$, with the largest gains concentrated in Step 2 (definitions, +0.30 points, that is 6.0\%, from $\theta$=~0.6 to 0.9), confirming $\theta$=~0.9 as the strongest configuration within the evaluated range. Second, the retry budget exhibits a striking non-linearity: a single retry yields negligible improvement, while the qualitative shift occurs at $r$=~2, where Step 2 recovers 1.25 points (25.0\%) as the candidate pool becomes large enough for the best-of-$r$ selection mechanism to escape low-quality basins; gains saturate beyond $r$=~2, and we adopt $r$=~3 as a conservative default. Third, our section-aware chunking strategy outperforms that of \citet{datla2025executable} by +0.16 points (3.2\%) overall, with the benefit concentrated entirely in early pipeline stages (+0.33 points, i.e., 6.6\%, at Step 1), confirming that downstream refinement cannot compensate for incoherent inputs. Fourth, conditional on an adequate retry budget, the choice of regeneration trigger (average-score vs. per-criterion) has no measurable effect on final quality, establishing the retry budget as the dominant control variable and trigger granularity as second-order.

\vspace{-0.3cm}
\section{Conclusion}\label{sec:conclusion}\vspace{-0.3cm}
We presented De~Jure, a fully automated, domain-agnostic pipeline that converts raw regulatory documents into structured, machine-readable rule sets by interleaving typed semantic extraction with a hierarchically ordered multi-stage LLM judge and an iterative repair mechanism. Metadata and definitions are corrected before rule units are evaluated, ensuring that each downstream stage operates on the best available upstream context and that errors do not silently propagate into rule decompositions. De~Jure achieves strong extraction quality on financial securities regulation and generalizes without modification to healthcare privacy and AI governance, maintaining consistently high performance across all model families and access paradigms. A downstream RAG-based evaluation demonstrates that De~Jure-grounded responses are strongly preferred over those from a strong prior approach, with the margin widening as retrieval depth increases, confirming that extraction fidelity translates directly into downstream utility. Ablation studies further validate the core design decisions: the retry budget is the dominant quality lever, and input chunking quality directly conditions extraction fidelity in early pipeline stages. Together, these results demonstrate that explicit, interpretable evaluation criteria can substitute for human annotation in annotation-scarce regulatory settings, and that iterative judge-guided refinement offers a scalable and auditable path toward regulation-grounded LLM alignment.

\bibliography{colm2026_conference}
\bibliographystyle{colm2026_conference}

\newpage
 
 \appendix
\section{Ablation Studies}
\label{sec:app_ablation}

We present four ablations targeting the core design decisions of our pipeline.
Unless stated otherwise, all experiments use the following defaults: maximum
regeneration retries $r{=}3$, acceptance threshold $\theta{=}0.9$,
\texttt{Llama-3.1-8B-Instruct} as the backbone, and HIPAA as the evaluation corpus.

\subsection{Impact of Acceptance Threshold}
\label{sec:ablation-threshold}

At each pipeline stage, an output is accepted only if its average quality score meets
or exceeds $\theta$; otherwise regeneration is triggered. We ablate
$\theta \in \{0.6, 0.7, 0.8, 0.9\}$ and report per-step quality in
Table~\ref{tab:ablation_threshold}.

\begin{table}[h]
\small
\setlength{\tabcolsep}{9pt}
\renewcommand{\arraystretch}{1.15}
\centering
\begin{tabular}{ccccc}
\toprule
$\theta$ & Step 1 & Step 2 & Step 3 & Total \\
\midrule
0.6                    & 4.86          & 4.12          & 4.47          & 4.48          \\
0.7                    & 4.86          & 4.21          & 4.49          & 4.52          \\
0.8                    & \textbf{4.92} & 4.31          & 4.58          & 4.61          \\
\textbf{0.9}           & \textbf{4.92} & \textbf{4.42} & \textbf{4.65} & \textbf{4.67} \\
\midrule
$\Delta$ (0.6$\to$0.9) & +0.06         & +0.30         & +0.18         & +0.19         \\
\bottomrule
\end{tabular}
\caption{Average quality score per pipeline step across acceptance thresholds $\theta$
(scale 1--5). Bold row denotes our default. $\Delta$ reports the absolute gain from
the weakest to the strongest threshold.}
\label{tab:ablation_threshold}
\end{table}

Quality improves monotonically with $\theta$, with the total average rising from 4.48
to 4.67. The $\Delta$ row exposes a structurally meaningful asymmetry: Step~2 captures
the dominant share of the gain (+0.30 points, i.e., 6\%), compared to only +0.06 points (1.2\%) for Step~1. Step~1
operates over well-scoped, bounded inputs where a first-pass extraction is typically
sufficient; Step~2 involves compositionally harder extraction decisions over loosely
structured definitional content, where the higher bar imposed by $\theta{=}0.9$
meaningfully increases the probability that a higher-quality candidate is surfaced
through regeneration. The consistent gains at $\theta{=}0.9$ across all stages, at
modest additional regeneration cost, confirm it as the optimal operating point.

\subsection{Impact of Maximum Regeneration Retries}
\label{sec:ablation-retries}

\begin{figure}[h]
\centering
\includegraphics[scale=0.6]{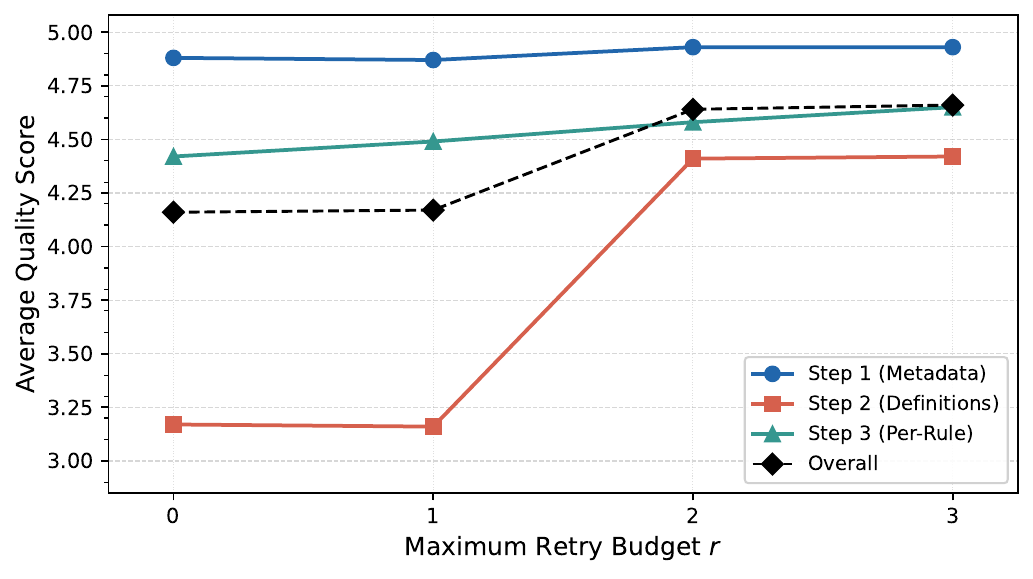}
\caption{Average quality score per pipeline step as a function of retry budget $r$
(scale 1--5; higher is better). Steps~1 and~3 remain largely flat throughout,
while Step~2 exhibits a sharp threshold effect: negligible gain from $r{=}0$ to
$r{=}1$, followed by a 1.25-point (25\%) recovery at $r{=}2$. The shaded region marks the
negligible-gain zone. All steps saturate beyond $r{=}2$.}
\label{fig:ablation_retries}
\end{figure}

The retry budget $r$ controls how many regeneration attempts a pipeline stage may make
before the best-scoring output is accepted. We ablate $r \in \{0, 1, 2, 3\}$, where
$r{=}0$ corresponds to a single-pass pipeline with no regeneration. Full per-step
scores across all values of $r$ are plotted in Figure~\ref{fig:ablation_retries}.

The results reveal a striking non-linearity concentrated entirely in Step~2.
Increasing $r$ from 0 to 1 yields a negligible total gain of 0.01 points. This near-zero first-retry
return is a diagnostic finding: the distributional tendencies that produced the
initial suboptimal output persists under minimally perturbed sampling, making a single
alternative draw unlikely to materially improve upon the failure.

The qualitative shift occurs at $r{=}2$, delivering a total gain of 0.47 points (9.4\%) over
$r{=}1$, driven almost entirely by Step~2 recovering 1.25 points (25\%). This threshold
behavior indicates that escaping the low-quality basin at Step~2 requires a minimum
candidate pool of three: the initial output establishes the failure mode, the first
retry explores a partial correction, and the second provides sufficient diversity for
the selection mechanism to identify a genuinely superior output. This is consistent
with the qualitative progression in Figure~\ref{fig:example}, where judge critiques
from earlier attempts provide increasingly targeted feedback that steers the model away
from prior failure modes rather than resampling from the same distribution. Steps~1
and~3 do not exhibit this pattern, confirming the non-linearity is specific to the
compositional complexity of Step~2. Beyond $r{=}2$, total gains saturate at +0.02,
and we adopt $r{=}3$ as the default, providing a conservative safety margin at
negligible additional cost.

\subsection{Impact of Chunking Strategy}
\label{sec:ablation-chunking}

Input chunk quality is a silent but consequential variable: each pipeline stage can
only reason over the regulatory context it receives, so structural artifacts introduced
at the chunking stage propagate forward. We compare our chunking strategy against
that of~\citet{datla2025executable} by routing both through an identical pipeline on
the same HIPAA subset.

\begin{table}[h]
\small
\setlength{\tabcolsep}{9pt}
\renewcommand{\arraystretch}{1.15}
\centering
\begin{tabular}{lcccc}
\toprule
\textbf{Chunking Strategy}   & Step 1        & Step 2        & Step 3        & Total         \\
\midrule
Ours                         & \textbf{4.97} & \textbf{4.47} & \textbf{4.60} & \textbf{4.68} \\
\citet{datla2025executable}  & 4.64          & 4.34          & 4.59          & 4.52          \\
\midrule
$\Delta$ (Ours $-$ comparison target) & +0.33         & +0.13         & +0.01         & +0.16         \\
\bottomrule
\end{tabular}
\caption{Average quality score per pipeline step for each chunking strategy on the
HIPAA subset (scale 1--5). $\Delta$ reports the absolute gain of our strategy.}
\label{tab:ablation_chunking}
\end{table}

The $\Delta$ row reveals a gain profile whose non-uniformity is diagnostically
informative. Step~1 captures nearly the entire benefit (+0.33 points, i. e., 6.6\% ): a chunk that cleanly
encapsulates a single regulatory provision allows precise, well-scoped extraction on  
the first attempt, whereas a poorly bounded chunk introduces structural ambiguity that
degrades initial quality before regeneration has any opportunity to help. Step~3, by
contrast, is virtually identical across strategies (+0.01), suggesting it operates
against a quality ceiling set by model capacity rather than input structure. Chunking
exerts its leverage exclusively in the early pipeline stages; downstream refinement
cannot compensate for incoherent inputs, it can only refine coherent ones.

\subsection{Regeneration Trigger Strategy}
\label{sec:ablation-trigger}

We evaluate two trigger strategies: \textbf{avg-trigger}, which regenerates if the
normalized average score across all criteria falls below $\theta{=}0.9$; and
\textbf{individual-trigger}, a strictly more conservative criterion that regenerates
if any single criterion falls below a raw score of 4. Both strategies are evaluated
under retry budgets $r \in \{1, 3\}$ to disentangle the effect of trigger policy from
retry budget.

\begin{table}[h]
\small
\setlength{\tabcolsep}{9pt}
\renewcommand{\arraystretch}{1.15}
\centering
\begin{tabular}{llcccc}
\toprule
\textbf{Max Retries ($r$)} & \textbf{Trigger} & Step 1 & Step 2 & Step 3 & Total \\
\midrule
\multirow{2}{*}{$r{=}3$} & Avg (default) & 4.92 & 4.42 & 4.65 & 4.66 \\
                          & Individual    & 4.92 & 4.44 & 4.63 & 4.66 \\
\midrule
\multirow{2}{*}{$r{=}1$} & Avg           & 4.87 & 3.16 & 4.49 & 4.18 \\
                          & Individual    & 4.87 & 3.21 & 4.43 & 4.17 \\
\midrule
$\Delta$ ($r$: 1$\to$3, Avg) &            & +0.05 & +1.26 & +0.16 & +0.48 \\
\bottomrule
\end{tabular}
\caption{Average quality score per pipeline step across trigger strategies and retry
budgets. $\Delta$ reports the gain from $r{=}1$ to $r{=}3$ under the
default avg-trigger.}
\label{tab:ablation_trigger}
\end{table}

Two findings stand out. First, the retry budget is the dominant factor by a wide
margin: increasing $r$ from 1 to 3 yields a total gain of 0.48 points (9.6\%), with Step~2
alone recovering 1.26 points (25.2\%). This asymmetry is structurally expected: Step~2 faces
the most compositionally demanding extraction task and routinely encounters diverse,
co-occurring failure modes that a single retry cannot simultaneously correct. The
budget of $r{=}3$ provides the minimum candidate pool necessary for the best-selection
mechanism to identify a materially superior output. Second, conditional on $r{=}3$,
both trigger strategies reach identical total quality (4.66), differing by at most 0.02 points (0.4\%)
on any individual step. With an adequate retry budget, the pipeline's best-of-$r$
selection naturally corrects per-criterion weaknesses without needing them to be
individually flagged. The individual-trigger's added conservatism imposes higher
regeneration and inference cost with zero measurable return. These results confirm
that trigger granularity is a second-order concern: retry budget is the primary
control variable.

\section{Judgment Criteria}
\label{sec:appx_judge_criteria}

De~Jure employs three specialized judges operating sequentially, each evaluating a
distinct semantic layer: section-level metadata (Judge~1), definitional content
(Judge~2), and individual rule units (Judge~3). Criteria within each judge are
designed to be \emph{orthogonal and collectively exhaustive} over the failure modes
specific to that layer. All criteria are scored on a 1--5 scale, and the per-stage
average determines whether the regeneration threshold $\theta$ is met.

\begin{table}[h]
\centering
\setlength{\tabcolsep}{10pt}
\renewcommand{\arraystretch}{1.2}
\begin{tabular}{lll}
\toprule
\textbf{Judge} & \textbf{Evaluates} & \textbf{Criteria} \\
\midrule
\multirow{6}{*}{Judge 1} & \multirow{6}{*}{Section Metadata}
  & Completeness \\
& & Fidelity to Source Text \\
& & Non-Hallucination \\
& & Title Quality \\
& & Precision of Citations and Dates \\
& & Meaningful Population of Optional Fields \\
\midrule
\multirow{5}{*}{Judge 2} & \multirow{5}{*}{Definitions}
  & Completeness \\
& & Fidelity to Source Text \\
& & Non-Hallucination \\
& & Precision and Formatting \\
& & Quality of Terms \\
\midrule
\multirow{8}{*}{Judge 3} & \multirow{8}{*}{Per-Rule Quality}
  & Completeness \\
& & Conciseness \\
& & Accuracy \\
& & Fidelity to Source Text \\
& & Neutrality \\
& & Consistency \\
& & Actionability \\
& & Non-Hallucination \\
\bottomrule
\end{tabular}
\caption{Judgment criteria grouped by pipeline stage. Each judge targets the failure
modes specific to its semantic layer: structural integrity (Judge~1), definitional
precision (Judge~2), and fine-grained rule correctness and utility (Judge~3).}
\label{tab:judgement-criteria}
\end{table}

\paragraph{Judge 1: Section Metadata.}
Criteria target both factual accuracy (citations, dates, titles) and extraction
completeness. \textit{Non-Hallucination} and \textit{Fidelity to Source Text} act as
hard correctness gates: a metadata extraction that distorts regulatory identifiers is
unusable regardless of structural quality. \textit{Meaningful Population of Optional
Fields} serves as a proxy for thoroughness beyond mandatory fields.

\paragraph{Judge 2: Definitions.}
This stage evaluates whether the extracted glossary is complete, source-faithful, and
well-formed. \textit{Precision and Formatting} is included because malformed
definitions degrade both retrieval and rule grounding downstream. \textit{Quality of
Terms} assesses whether extracted terms are genuine regulatory primitives rather than
incidental phrases, a distinction requiring semantic judgment beyond surface copying.

\paragraph{Judge 3: Per-Rule Quality.}
The most demanding stage, assessing each rule unit across eight criteria organized into
three functional clusters: \textit{Completeness}, \textit{Accuracy}, and
\textit{Actionability} assess whether the rule captures the full operative content of
the source provision; \textit{Fidelity to Source Text}, \textit{Neutrality}, and
\textit{Non-Hallucination} form a faithfulness cluster ensuring the extraction neither
omits nor introduces content; and \textit{Conciseness} and \textit{Consistency} assess
structural quality, as redundant or contradictory rules impose unnecessary burden on
downstream retrieval and synthesis.

\subsection{Pairwise Judge Criteria for Downstream Compliance QA}
\label{appx:ds_judge_criteria}

While the three judges above assess intrinsic extraction quality, Experiment~3 
(Section~\ref{sec:downstream-eval}) requires a separate instrument to measure 
downstream utility. A pairwise LLM judge compares RAG responses produced from 
De~Jure extractions against those of \citet{datla2025executable} across six criteria. 
\textit{Completeness} captures whether all aspects of the question are addressed; 
\textit{Factual Grounding} penalizes claims not traceable to the retrieved rule set; 
\textit{Handling Ambiguity} assesses whether the response correctly distinguishes 
mandates, permissions, and unresolved provisions rather than forcing false certainty; 
\textit{Practical Actionability} measures whether regulatory language is translated 
into concrete guidance rather than accurate but inert quotation; 
\textit{Regulatory Precision} evaluates correct reflection of scope, conditions, and 
exceptions; and \textit{Overall Preference} provides a holistic judgment integrating 
all five dimensions, reported separately to surface trade-offs not captured by any 
single criterion.

These criteria collectively operationalize \emph{downstream utility}. In regulatory 
compliance, a well-formed extraction has no value unless it enables a system to answer 
questions that are complete, grounded, unambiguous, and actionable. Each criterion 
directly targets a failure mode that poor rule extraction introduces into the RAG 
pipeline: incomplete rules truncate responses, hallucinated content propagates into 
assertions, imprecise definitions blur scope, and incoherent structure impedes 
actionable synthesis. Evaluating at this level provides a direct, task-grounded 
measure of whether extraction fidelity translates into practical compliance support.

\clearpage

\section{Pipeline Success Example: Judgment and Repair in Practice}
\label{appx:example}

This section walks through a concrete example of judge-guided repair. A structurally
complete but semantically flawed extraction is identified by the judge, corrected in
a single repair step, and re-evaluated to a passing score. The example also shows
that the judge scores each criterion independently: it penalizes only the fields that
are wrong and preserves high scores on fields that are already correct.

\begin{figure*}[h]
\centering
\includegraphics[width=\linewidth]{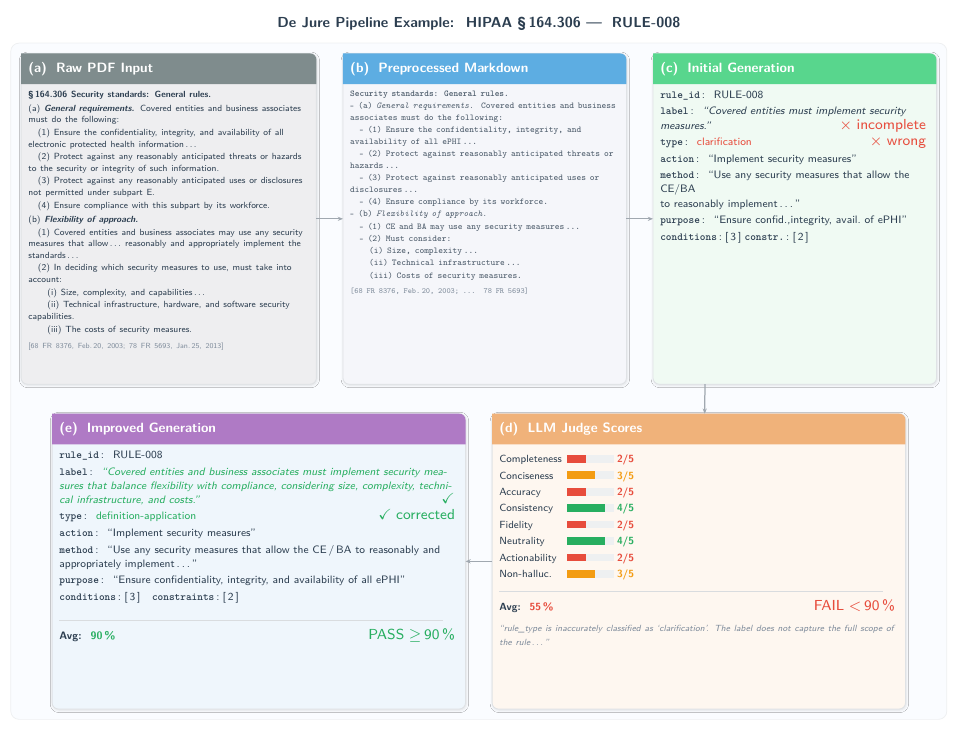}
\caption{De~Jure applied to HIPAA \S\,164.306. Panels (c)→(d)→(e) read left-to-right across the bottom row. Panels~(a)--(b) show the raw PDF
source and its pre-processed Markdown. Panel~(c) shows the initial extraction with
two field-level defects: an incomplete label and a misclassified rule type.
Panel~(d) shows the judge evaluation (avg.\ 0.55, \textsc{fail}) with per-criterion
scores and targeted critiques. Panel~(e) shows the corrected extraction after a
single repair iteration (avg.\ 0.90, \textsc{pass}), with only deficient fields
revised and all others preserved.}
\label{fig:example}
\end{figure*}

Figure~\ref{fig:example} traces RULE-008 from HIPAA \S\,164.306 through the full
pipeline. Panels~(a)--(b) show the raw source and its pre-processed Markdown. The
four blocks below correspond to: initial extraction, judge evaluation of the initial
extraction, corrected extraction, and judge re-evaluation of the corrected extraction.

\medskip
\noindent\textbf{Block~1: Initial Extraction.}
The initial extraction is structurally complete as per the extraction schema (see Appendix~\ref{appx:schema}), and free of hallucinations. Targets,
conditions, and the verbatim span are all correct. However, two fields are defective.
The label is too brief: \textit{"Covered entities must implement security measures"}
omits the multi-factor balancing that is central to the provision. The
\texttt{rule\_type} is misclassified as \texttt{clarification} instead of
\texttt{definition-application}, silently reversing the compliance semantics. 

\begin{tcolorbox}[
  enhanced, breakable,
  colback=orange!3, colframe=orange!45, boxrule=0.55pt, arc=3pt,
  left=5pt, right=5pt, top=3pt, bottom=3pt,
  attach boxed title to top left={yshift=-2.5mm, xshift=4mm},
  boxed title style={enhanced, colback=orange!35, colframe=orange!45,
    boxrule=0.4pt, arc=2pt, left=4pt, right=4pt, top=2pt, bottom=2pt},
  title={\textbf{\small\sffamily Block~1: Initial Extraction (RULE-008)}},
  fontupper=\fontsize{5.8}{7.5}\selectfont\ttfamily,
]
\bigskip
\{\\
\hspace{3mm}"rule\_id": "RULE-008",\\
\hspace{3mm}"label": "Covered entities must implement security measures.",\\
\hspace{3mm}"rule\_type": \{ "type": \textbf{"clarification"}, "other\_label": "implementation-guidance" \},\\
\hspace{3mm}"targets": [ \{ "role": "covered\_entity" \}, \{ "role": "business\_associate" \} ],\\
\hspace{3mm}"statement": \{\\
\hspace{6mm}"action": "Implement security measures",\\
\hspace{6mm}"action\_object": "security measures",\\
\hspace{6mm}"method": "Use any security measures that allow the covered entity or business\\
\hspace{12mm}associate to reasonably and appropriately implement the standards and\\
\hspace{12mm}implementation specifications as specified in this subpart, taking into\\
\hspace{12mm}account the size, complexity, and capabilities of the covered entity or\\
\hspace{12mm}business associate, technical infrastructure, hardware, and software\\
\hspace{12mm}security capabilities, and costs of security measures.",\\
\hspace{6mm}"constraints": [\\
\hspace{9mm}\{ "text": "Reasonably anticipated threats or hazards to the security or\\
\hspace{15mm}integrity of electronic protected health information" \},\\
\hspace{9mm}\{ "text": "Reasonably anticipated uses or disclosures of electronic protected\\
\hspace{15mm}health information not permitted or required under subpart E" \}\\
\hspace{6mm}],\\
\hspace{6mm}"conditions": [\\
\hspace{9mm}\{ "trigger": "size, complexity, and capabilities of the covered entity or business associate" \},\\
\hspace{9mm}\{ "trigger": "technical infrastructure, hardware, and software security capabilities" \},\\
\hspace{9mm}\{ "trigger": "costs of security measures" \}\\
\hspace{6mm}],\\
\hspace{6mm}"exceptions": [],\\
\hspace{6mm}"penalties\_or\_consequences": null,\\
\hspace{6mm}"purpose": "Ensure the confidentiality, integrity, and availability of all\\
\hspace{12mm}electronic protected health information",\\
\hspace{6mm}"verbatim": "Covered entities and business associates may use any security\\
\hspace{12mm}measures that allow the covered entity or business associate to reasonably\\
\hspace{12mm}and appropriately implement the standards and implementation specifications\\
\hspace{12mm}as specified in this subpart."\\
\hspace{3mm}\},\\
\hspace{3mm}"citations": \{ "text": "164.306: Security standards: General rules.\\
\hspace{12mm}[68 FR 8376, Feb. 20, 2003; 68 FR 17153, Apr. 8, 2003; 78 FR 5693, Jan. 25, 2013]" \},\\
\hspace{3mm}"examples": []\\
\}
\end{tcolorbox}

\medskip
\noindent\textbf{Block~2: Judge Evaluation (Fail, 0.55).}
The judge returns a normalized average score of 0.55 and flags the extraction as \textsc{fail}. The
critique is targeted: it assigns low scores to \textit{Completeness},
\textit{Accuracy}, \textit{Fidelity}, and \textit{Actionability}, which are the
affected criteria, while keeping high scores on \textit{Consistency} and
\textit{Neutrality}, which are correct. This field-level signal is what makes the
repair step precise. The feedback for each criteria is provided to guide the regeneration in a direction to improve the score on the respected criteria.

\begin{tcolorbox}[
  enhanced, breakable,
  colback=orange!4, colframe=orange!55!black, boxrule=0.55pt, arc=3pt,
  left=5pt, right=5pt, top=3pt, bottom=3pt,
  attach boxed title to top left={yshift=-2.5mm, xshift=4mm},
  boxed title style={enhanced, colback=orange!60!black, colframe=orange!55!black,
    boxrule=0.4pt, arc=2pt, left=4pt, right=4pt, top=2pt, bottom=2pt},
  title={\textbf{\small\sffamily\color{white} Block~2: Judge Evaluation of Initial Extraction
    \textnormal{\color{white!80!orange} $\mid$} Avg.\ 0.55 (\textsc{fail})}},
  fontupper=\fontsize{5.8}{7.5}\selectfont\ttfamily,
]
\bigskip
"Completeness":   \{ \textbf{"Score": 2}, "Justification": "Label incompletely summarizes the rule,\\
\hspace{4mm}omitting the multi-factor balancing requirement central to the provision.\\
\hspace{4mm}rule\_type is misclassified; should be definition-application." \},\\
"Conciseness":    \{ \textbf{"Score": 3}, "Justification": "Label is brief but sacrifices accuracy\\
\hspace{4mm}for brevity; the scope of the rule is inadequately represented." \},\\
"Accuracy":       \{ \textbf{"Score": 2}, "Justification": "rule\_type classified as clarification\\
\hspace{4mm}instead of definition-application, misrepresenting the rule's nature." \},\\
"Consistency":    \{ \textbf{"Score": 4}, "Justification": "Targets correctly identified as\\
\hspace{4mm}covered\_entity and business\_associate, consistent with source." \},\\
"Fidelity":       \{ \textbf{"Score": 2}, "Justification": "Statement omits the full balancing\\
\hspace{4mm}framework including both targets; significant omission relative to source." \},\\
"Neutrality":     \{ \textbf{"Score": 4}, "Justification": "Label and statement are neutrally\\
\hspace{4mm}presented; no interpretive bias introduced." \},\\
"Actionability":  \{ \textbf{"Score": 2}, "Justification": "Statement fails to convey what both\\
\hspace{4mm}covered entities and business associates must do; insufficiently operational." \},\\
"Non\_hallucination": \{ \textbf{"Score": 3}, "Justification": "No fabricated content, but\\
\hspace{4mm}significant omissions reduce faithfulness to the source provision." \}\\
\\
\textbf{Average Normalized Score: 0.55 -- FAIL -- Regeneration triggered.}
\end{tcolorbox}

\medskip
\noindent\textbf{Block~3: Corrected Extraction.}
After one repair iteration, only the two flagged fields change. The label is expanded
to capture the multi-factor balancing framework. The \texttt{rule\_type} is corrected
to \texttt{definition-application}. All other fields, including conditions,
constraints, verbatim span, and citations, remain exactly as before. The repair
targets specific deficiencies rather than regenerating the rule from scratch.

\begin{tcolorbox}[
  enhanced, breakable,
  colback=green!3, colframe=green!45!black, boxrule=0.55pt, arc=3pt,
  left=5pt, right=5pt, top=3pt, bottom=3pt,
  attach boxed title to top left={yshift=-2.5mm, xshift=4mm},
  boxed title style={enhanced, colback=green!45!black, colframe=green!45!black,
    boxrule=0.4pt, arc=2pt, left=4pt, right=4pt, top=2pt, bottom=2pt},
  title={\textbf{\small\sffamily\color{white} Block~3: Corrected Extraction (RULE-008)}},
  fontupper=\fontsize{5.8}{7.5}\selectfont\ttfamily,
]
\bigskip
\{\\
\hspace{3mm}"rule\_id": "RULE-008",\\
\hspace{3mm}"label": \textbf{"Covered entities and business associates must implement security}\\
\hspace{12mm}\textbf{measures that balance flexibility with compliance, considering size,}\\
\hspace{12mm}\textbf{complexity, technical infrastructure, and costs."},\\
\hspace{3mm}"rule\_type": \{ \textbf{"type": "definition-application"}, "other\_label": "implementation-guidance" \},\\
\hspace{3mm}"targets": [ \{ "role": "covered\_entity" \}, \{ "role": "business\_associate" \} ],\\
\hspace{3mm}"statement": \{\\
\hspace{6mm}"action": "Implement security measures",\\
\hspace{6mm}"action\_object": "security measures",\\
\hspace{6mm}"method": "Use any security measures that allow the covered entity or business\\
\hspace{12mm}associate to reasonably and appropriately implement the standards and\\
\hspace{12mm}implementation specifications as specified in this subpart, taking into\\
\hspace{12mm}account the size, complexity, and capabilities of the covered entity or\\
\hspace{12mm}business associate, technical infrastructure, hardware, and software\\
\hspace{12mm}security capabilities, and costs of security measures.",\\
\hspace{6mm}"constraints": [\\
\hspace{9mm}\{ "text": "Reasonably anticipated threats or hazards to the security or\\
\hspace{15mm}integrity of electronic protected health information" \},\\
\hspace{9mm}\{ "text": "Reasonably anticipated uses or disclosures of electronic protected\\
\hspace{15mm}health information not permitted or required under subpart E" \}\\
\hspace{6mm}],\\
\hspace{6mm}"conditions": [\\
\hspace{9mm}\{ "trigger": "size, complexity, and capabilities of the covered entity or business associate" \},\\
\hspace{9mm}\{ "trigger": "technical infrastructure, hardware, and software security capabilities" \},\\
\hspace{9mm}\{ "trigger": "costs of security measures" \}\\
\hspace{6mm}],\\
\hspace{6mm}"exceptions": [],\\
\hspace{6mm}"penalties\_or\_consequences": null,\\
\hspace{6mm}"purpose": "Ensure the confidentiality, integrity, and availability of all\\
\hspace{12mm}electronic protected health information",\\
\hspace{6mm}"verbatim": "Covered entities and business associates may use any security\\
\hspace{12mm}measures that allow the covered entity or business associate to reasonably\\
\hspace{12mm}and appropriately implement the standards and implementation specifications\\
\hspace{12mm}as specified in this subpart."\\
\hspace{3mm}\},\\
\hspace{3mm}"citations": \{ "text": "164.306: Security standards: General rules.\\
\hspace{12mm}[68 FR 8376, Feb. 20, 2003; 68 FR 17153, Apr. 8, 2003; 78 FR 5693, Jan. 25, 2013]" \},\\
\hspace{3mm}"examples": []\\
\}
\end{tcolorbox}

\medskip
\noindent\textbf{Block~4: Judge Re-evaluation (Pass, 0.90).}
The judge re-evaluates the corrected extraction and returns a normalized average score of 0.90. All
four previously low criteria recover: \textit{Completeness} and \textit{Accuracy}
reach 5, reflecting the corrected rule type and the richer label. \textit{Actionability}
and \textit{Fidelity} similarly improve. \textit{Conciseness} drops slightly, as the
longer label trades brevity for coverage. Criteria that were already correct retain
their scores, confirming that the judge does not penalize correct fields for the
failures of others. The score meets the threshold $\theta = 0.90$, and the repair
loop halts.

\begin{tcolorbox}[
  enhanced, breakable,
  colback=green!3, colframe=green!45!black, boxrule=0.55pt, arc=3pt,
  left=5pt, right=5pt, top=3pt, bottom=3pt,
  attach boxed title to top left={yshift=-2.5mm, xshift=4mm},
  boxed title style={enhanced, colback=green!45!black, colframe=green!45!black,
    boxrule=0.4pt, arc=2pt, left=4pt, right=4pt, top=2pt, bottom=2pt},
  title={\textbf{\small\sffamily\color{white} Block~4: Judge Evaluation of Corrected Extraction
    \textnormal{\color{white!80!green} $\mid$} Avg.\ 0.90 (\textsc{pass})}},
  fontupper=\fontsize{5.8}{7.5}\selectfont\ttfamily,
]
\bigskip
"Completeness":   \{ \textbf{"Score": 5}, "Justification": "Label now comprehensively captures\\
\hspace{4mm}the multi-factor balancing requirement. All required fields populated correctly.\\
\hspace{4mm}rule\_type corrected to definition-application." \},\\
"Conciseness":    \{ \textbf{"Score": 4}, "Justification": "Label is necessarily detailed to\\
\hspace{4mm}reflect the provision's scope; slight verbosity is justified given the\\
\hspace{4mm}complexity of the rule." \},\\
"Accuracy":       \{ \textbf{"Score": 5}, "Justification": "rule\_type correctly classified as\\
\hspace{4mm}definition-application; accurately represents the provision's regulatory\\
\hspace{4mm}function." \},\\
"Consistency":    \{ \textbf{"Score": 4}, "Justification": "Targets consistently and correctly\\
\hspace{4mm}identify both covered\_entity and business\_associate throughout." \},\\
"Fidelity":       \{ \textbf{"Score": 4}, "Justification": "Statement closely reflects the source\\
\hspace{4mm}text; label now captures both targets and the balancing framework.\\
\hspace{4mm}Minor omission: flexibility aspect could be made more explicit." \},\\
"Neutrality":     \{ \textbf{"Score": 5}, "Justification": "Label and statement remain neutrally\\
\hspace{4mm}framed with no interpretive bias introduced." \},\\
"Actionability":  \{ \textbf{"Score": 4}, "Justification": "Statement now provides operationally\\
\hspace{4mm}useful guidance for both covered entities and business associates, though\\
\hspace{4mm}method field is dense and could be more concisely structured." \},\\
"Non\_hallucination": \{ \textbf{"Score": 5}, "Justification": "All extracted content is\\
\hspace{4mm}grounded in the source text; no fabricated or inferred content introduced." \}\\
\\
\textbf{Average Normalized Score: 0.90 -- PASS -- Repair loop terminates.}
\end{tcolorbox}

\medskip
\noindent The jump of the normalized averaged score from 0.55 to 0.90 was driven by two field corrections, with everything else unchanged. This pattern holds across all corpora and aligns with
the quantitative results in Section~\ref{sec:ablation-retries}.

\medskip
\noindent This example illustrates two key properties of the judge. First, it
reliably flags weak extractions: the low scores in Block~2 are not generic
penalties but precise signals tied to specific fields, paired with actionable
feedback that directly guides the repair. Second, it recognizes improvement:
once those fields are corrected, the judge awards high scores to the updated
output without penalizing fields that were already correct. These two properties
are what make automated iterative repair practical. The judge distinguishes poor
extractions from good ones, and its feedback is specific enough to drive targeted
corrections rather than blind regeneration. This allows the extraction and verification to operate as mutually reinforcing stages and drive De Jure's self-correcting nature.

\newpage
\section{De~Jure Pipeline: Algorithm Psuedo Code}
\label{appx:alg_diag}

Algorithm~\ref{alg:pipeline} provides a complete procedural specification of the
De~Jure pipeline, complementing the stage-by-stage description in
Section~\ref{sec:main_method}. 

\begin{algorithm}[h]
\SetNoFillComment
\caption{De~Jure: Regulatory Rule Extraction with Judgment and Repair}
\label{alg:pipeline}

\Input{Document $D$ (PDF\,/\,HTML)}
\Output{Structured rule set $\mathcal{R}$}
\BlankLine

\tcc{Stage 1: Pre-processing}
$M \gets$ \Convert{$D$}\;
$M \gets$ \Clean{$M$}\;
$\mathcal{S} \gets$ \Split{$M$} \tcp*{split on section markers}
\ForEach{section $s \in \mathcal{S}$}{
  \Index{$s$}\;
  Attach metadata and \Hash{$D$} to $s$\;
}
\BlankLine

\tcc{Stage 2: Generation, Judgment, and Selective Repair}
$\mathcal{R} \gets \emptyset$\;
$r \gets 3$ \tcp*{max regeneration attempts per stage}
$\theta \gets 0.90$ \tcp*{avg-score acceptance threshold}
stages $\gets$ [(metadata, 6), (definitions, 5), (rule\_units, 8)]\;
\BlankLine

\ForEach{section $s \in \mathcal{S}$}{
  $e \gets$ \Generate{$s$, \texttt{JSONSchema}}\;
  \lIf{$e =$ \texttt{null}}{\textbf{continue} \tcp*{non-actionable section}}
  \BlankLine
  \ForEach{$(name,\, n_c) \in$ stages}{
    $\hat{e} \gets e$\;
    $\hat{\mu} \gets 0$ \tcp*{best score seen for this stage}
    \For{$t \gets 1$ \KwTo $r$}{
      $scores, critique \gets$ \Judge{$e$, $name$, $n_c$ criteria}\;
      $\mu \gets \mathrm{avg}(scores)$\;
      \If{$\mu > \hat{\mu}$}{
        $\hat{e} \gets e$\;
        $\hat{\mu} \gets \mu$\;
      }
      \lIf{$\mu \geq \theta$}{\textbf{break} \tcp*{threshold met}}
      $e \gets$ \Generate{$s$, $e$, $scores$, $critique$, \texttt{JSONSchema}}\;
    }
    $e \gets \hat{e}$ \tcp*{commit best generation for this stage}
  }
  $\mathcal{R} \gets \mathcal{R} \cup \{e\}$\;
}
\BlankLine
\Return $\mathcal{R}$\;
\end{algorithm}

The total LLM call budget per section is bounded by $1 + 3 \times r$ in the worst
case (one initial generation plus at most $r$ repair attempts across each of the three
stages), and reduces to $1 + 3$ in the best case where all stages pass on the first
judgment.

\newpage
\section{Extraction Quality Experiment: Full Results}
\label{appx:extraction_quality_results}

This section provides further details and extended results for the experiments presented in Section~\ref{sec:exp1}. 

Table \ref{tab:extraction_step1_criteria} presents the complete Judge 1 metadata criteria scores  for Experiment 1 across the four different models evaluated on SEC corpus. One observation that stands out is that Non-Hallucination and Fidelity to source text is uniformly perfect (5.00) across all models, confirming that schema-constrained extraction eliminates factual fabrication as a failure mode regardless of model family or access paradigm.

\begin{table*}[h]
\centering
{\footnotesize
\begin{tabular*}{\textwidth}{@{\extracolsep{\fill}}llcccccc}
\toprule
\textbf{Model} & \textbf{Completeness} & \textbf{Fidelity} & \textbf{Non-Halluc.} & \textbf{Title Quality} & \textbf{Citation \& Date} & \textbf{Opt.} & \textbf{Avg}\\
\midrule
Claude-3.5-Sonnet
    & \textbf{4.95} & \textbf{5.00} & \textbf{5.00} & \textbf{5.00} & \textbf{5.00} & \textbf{5.00} & \textit{\textbf{4.99}} \\
Llama-3.1-8B
    & 4.84 & \textbf{5.00} & \textbf{5.00} & 4.95 & 4.89 & 4.92 & \textit{4.93} \\
GPT-5-mini
    & \textbf{4.95} & \textbf{5.00} & \textbf{5.00} & \textbf{5.00} & \textbf{5.00} & 4.97 & \textbf{4.99}\\
Qwen3-VL-8B
    & 4.85 & \textbf{5.00} & \textbf{5.00} & \textbf{5.00} & 4.97 & 4.77 & \textit{\textbf{4.99}} \\
\bottomrule
\end{tabular*}
}
\caption{Full per-criteria scores on the SEC Investment Advisers Act corpus for \textbf{Judge 1} (Metadata) criteria (scale 1--5; higher is better). Abbreviations: Non-Halluc. = Non-Hallucination.; Opt. = Optional Field Population; Avg = Average Score}
\label{tab:extraction_step1_criteria}
\end{table*}

Table \ref{tab:extraction_step2_criteria} presents the complete Judge 2 definition criteria scores  for Experiment 1 across the four different models evaluated on SEC corpus. Non-Hallucination remains uniformly perfect (5.00), indicating strong faithfulness under the structured extraction setup. Compared to metadata, greater variation is observed in Completeness and Precision, reflecting the higher compositional complexity of definitions, with \textbf{Qwen3-VL-8B} achieving the highest overall score (4.87) while all models remain closely clustered.

\begin{table*}[h]
\centering
{\footnotesize
\begin{tabular*}{\textwidth}{@{\extracolsep{\fill}}llccccc}
\toprule
\textbf{Model} & \textbf{Completeness} & \textbf{Fidelity} & \textbf{Non-Halluc.} & \textbf{Precision} & \textbf{Term Quality} & \textbf{Avg}\\
\midrule
Claude-3.5-Sonnet
    & 4.59 & 4.90 & \textbf{5.00} & 4.77 & 4.90 & \textit{4.83} \\
Llama-3.1-8B
    & 4.50 & 4.76 & \textbf{5.00} & 4.66 & 4.84 & \textit{4.75} \\
GPT-5-mini
    & 4.54 & \textbf{4.95} & \textbf{5.00} & 4.69 & \textbf{4.95} & \textit{4.83}\\
Qwen3-VL-8B
    & \textbf{4.69} & 4.90 & \textbf{5.00} & \textbf{4.85} & 4.92 & \textit{\textbf{4.87}} \\
\bottomrule
\end{tabular*}
}
\caption{Full per-criteria scores on the SEC Investment Advisers Act corpus for \textbf{Judge 2} (Definitions) criteria (scale 1–5; higher is better). Abbreviations: Non-Halluc. = Non-Hallucination; Precision = Precision \& Formatting.}
\label{tab:extraction_step2_criteria}
\end{table*}

Table \ref{tab:extraction_step3_criteria} presents the complete Judge 3 per-rule criteria scores  for Experiment 1 across the four different models evaluated on SEC corpus. All the models again obtain perfect scores for Non-Hallucination as well as Neutrality, indicating strong faithfulness under the structured extraction setup. Performance differences are most pronounced in Accuracy, Consistency, and Fidelity, where \textbf{GPT-5-mini} leads overall (4.75 \textit{Avg}), while open-source models remain competitive, with \textbf{Qwen3-VL-8B} achieving comparable scores (4.64), highlighting the robustness of the pipeline across model families.

\begin{table*}[h]
\centering
{\footnotesize
\begin{tabular*}{\textwidth}{@{\extracolsep{\fill}}llcccccccc}
\toprule
\textbf{Model} & \textbf{Comp.} & \textbf{Conc.} & \textbf{Accuracy} & \textbf{Cons.} & \textbf{Fidelity} & \textbf{Neut.} & \textbf{Actionab.} & \textbf{Non-Halluc.} & \textbf{Avg} \\
\midrule
Claude-3.5-Sonnet
    & 4.11 & \textbf{4.68} & 4.80 & 4.95 & 4.43 & \textbf{5.00} & 4.42 & \textbf{5.00} & \textit{4.67} \\
Llama-3.1-8B
    & 4.06 & 4.40 & 4.57 & 4.72 & 4.26 & \textbf{5.00} & 4.25 & \textbf{5.00} & \textit{4.53} \\
GPT-5-mini
    & \textbf{4.27} & 4.49 & \textbf{4.95} & \textbf{4.98} & \textbf{4.74} & \textbf{5.00} & \textbf{4.54} & \textbf{5.00} & \textit{\textbf{4.75}} \\
Qwen3-VL-8B
    & 4.10 & 4.48 & 4.81 & 4.93 & 4.43 & \textbf{5.00} & 4.36 & \textbf{5.00} & \textit{4.64} \\
\bottomrule
\end{tabular*}
}
\caption{Full per-criteria scores on the SEC Investment Advisers Act corpus for \textbf{Judge 3} (Per-Rule) criteria (scale 1–5; higher is better). Abbreviations: Comp. = Completeness; Conc. = Conciseness; Cons. = Consistency; Neut. = Neutrality; Actionab. = Actionability; Non-Halluc. = Non-Hallucination.}
\label{tab:extraction_step3_criteria}
\end{table*}

\newpage
\section{Domain Generalization Experiment: Full Results}
\label{appx:gen_results}

This section provides further details and extended results for the experiments presented in Section~\ref{sec:exp2}. \vspace{-0.1cm}

Table~\ref{tab:gen_step1_criteria} presents the complete Experiment 2 results for Judge~1 metadata criteria across SEC, EU AI Act, and HIPAA. \textbf{Qwen3-VL-8B} and \textbf{GPT-5-mini} perform comparably overall on metadata extraction, with only small differences across most criteria. Notably, the EU AI Act is consistently lower than the other two datasets on both \textbf{Completeness} and \textbf{Citation \& Date}, indicating that metadata recovery is more challenging in this corpus. \vspace{-0.1cm}

\begin{table*}[h]
\centering
{\footnotesize
\begin{tabular*}{\textwidth}{@{\extracolsep{\fill}}llcccccc}
\toprule
\textbf{Dataset} & \textbf{Model} & \textbf{Completeness} & \textbf{Fidelity} & \textbf{Non-Halluc.} & \textbf{Title Quality} & \textbf{Citation \& Date} & \textbf{Opt.} \\
\midrule
\multirow{2}{*}{SEC}
	& Qwen3-VL-8B & 4.85 & 5.00 & 5.00 & 5.00 & 4.97 & 4.77 \\
	& GPT-5-mini  & 4.95 & 5.00 & 5.00 & 5.00 & 5.00 & 4.97 \\
\cmidrule(lr){1-8}
\multirow{2}{*}{EU AI Act}
	& Qwen3-VL-8B & 3.89 & 5.00 & 5.00 & 5.00 & 4.67 & 4.33 \\
	& GPT-5-mini  & 4.00 & 5.00 & 5.00 & 5.00 & 4.78 & 4.56 \\
\cmidrule(lr){1-8}
\multirow{2}{*}{HIPAA}
	& Qwen3-VL-8B & 4.85 & 4.95 & 4.93 & 5.00 & 4.90 & 4.76 \\
	& GPT-5-mini  & 4.85 & 5.00 & 5.00 & 5.00 & 4.95 & 4.76 \\
\bottomrule
\end{tabular*}
}
\caption{Generalization across three regulatory corpora for \textbf{Judge 1} (Metadata) criteria (scale 1--5; higher is better). Abbreviations: Non-Halluc. = Non-Hallucination.; Opt. = Optional Field Population}
\label{tab:gen_step1_criteria}
\end{table*}

Table~\ref{tab:gen_step2_criteria} presents the complete Experiment 2 results for Judge~2 definition criteria across SEC, EU AI Act, and HIPAA. \textbf{GPT-5-mini} underperforms on definition \textbf{Completeness} and \textbf{Precision}, with the largest drop observed on HIPAA. More broadly, HIPAA appears to be the most difficult corpus for definition extraction, with both models showing lower scores than on SEC and EU AI Act in key definition quality dimensions. \vspace{-0.1cm}

\begin{table*}[h]
\centering
{\footnotesize
\begin{tabular*}{\textwidth}{@{\extracolsep{\fill}}llccccc}
\toprule
\textbf{Dataset} & \textbf{Model} & \textbf{Completeness} & \textbf{Fidelity} & \textbf{Non-Halluc.} & \textbf{Precision} & \textbf{Term Quality} \\
\midrule
\multirow{2}{*}{SEC}
	& Qwen3-VL-8B & 4.69 & 4.90 & 5.00 & 4.85 & 4.92 \\
	& GPT-5-mini  & 4.54 & 4.95 & 5.00 & 4.69 & 4.95 \\
\cmidrule(lr){1-7}
\multirow{2}{*}{EU AI Act}
	& Qwen3-VL-8B & 4.67 & 4.89 & 5.00 & 4.67 & 4.89 \\
	& GPT-5-mini  & 4.44 & 4.78 & 5.00 & 4.44 & 4.78 \\
\cmidrule(lr){1-7}
\multirow{2}{*}{HIPAA}
	& Qwen3-VL-8B & 4.51 & 4.71 & 5.00 & 4.63 & 4.68 \\
	& GPT-5-mini  & 4.20 & 4.76 & 5.00 & 4.32 & 4.78 \\
\bottomrule
\end{tabular*}
}
\caption{Generalization across three regulatory corpora for \textbf{Judge 2} (Definitions) criteria (scale 1--5; higher is better). Abbreviations: Non-Halluc. = Non-Hallucination; Precision = Precision \& Formatting.}
\label{tab:gen_step2_criteria}
\end{table*}

Table~\ref{tab:gen_step3_criteria} presents the complete Experiment 2 results for Judge~3 per-rule criteria across SEC, EU AI Act, and HIPAA. \textbf{Qwen3-VL-8B} underperforms relative to \textbf{GPT-5-mini} on rule \textbf{Completeness}, \textbf{Fidelity}, and \textbf{Actionability} across datasets. At the same time, both models perform strongly on \textbf{Accuracy}, \textbf{Consistency}, and \textbf{Neutrality}. Finally, consistent with Tables~\ref{tab:gen_step1_criteria} and \ref{tab:gen_step2_criteria}, hallucination remains minimal across all three judges, with near-ceiling non-hallucination scores throughout. \vspace{-0.1cm}

\begin{table*}[h]
\centering
{\footnotesize
\begin{tabular*}{\textwidth}{@{\extracolsep{\fill}}llcccccccc}
\toprule
\textbf{Dataset} & \textbf{Model} & \textbf{Comp.} & \textbf{Conc.} & \textbf{Accuracy} & \textbf{Cons.} & \textbf{Fidelity} & \textbf{Neut.} & \textbf{Actionab.} & \textbf{Non-Halluc.} \\
\midrule
\multirow{2}{*}{SEC}
	& Qwen3-VL-8B & 4.10 & 4.48 & 4.81 & 4.93 & 4.43 & 5.00 & 4.36 & 5.00 \\
	& GPT-5-mini  & 4.27 & 4.49 & 4.95 & 4.98 & 4.74 & 5.00 & 4.54 & 5.00 \\
\cmidrule(lr){1-10}
\multirow{2}{*}{EU AI Act}
	& Qwen3-VL-8B & 4.28 & 4.41 & 4.90 & 4.93 & 4.51 & 5.00 & 4.19 & 5.00 \\
	& GPT-5-mini  & 4.42 & 4.35 & 4.95 & 4.98 & 4.65 & 5.00 & 4.34 & 5.00 \\
\cmidrule(lr){1-10}
\multirow{2}{*}{HIPAA}
	& Qwen3-VL-8B & 4.15 & 4.52 & 4.84 & 4.96 & 4.50 & 5.00 & 4.32 & 4.99 \\
	& GPT-5-mini  & 4.31 & 4.52 & 4.95 & 4.99 & 4.76 & 5.00 & 4.46 & 5.00 \\
\bottomrule
\end{tabular*}
}
\caption{Generalization across three regulatory corpora for \textbf{Judge 3} (Per-Rule) criteria (scale 1--5; higher is better). Abbreviations: Comp. = Completeness; Conc. = Conciseness; Cons. = Consistency; Neut. = Neutrality; Actionab. = Actionability; Non-Halluc. = Non-Hallucination.}
\label{tab:gen_step3_criteria}
\end{table*}

\newpage
\section{Prompts}
\label{appx:prompts}

De~Jure uses seven prompts organized into three functional groups: an initial
extraction prompt that generates the structured rule representation from raw text
(Section~\ref{appx:extraction_prompt}), three stage-specific regeneration prompts
that incorporate judge feedback to correct failing extractions
(Section~\ref{appx:regen_prompts}), and three judge prompts that produce the
per-criterion scores and natural-language critiques driving the repair loop
(Section~\ref{appx:judge_prompts}). All prompts are fully domain-agnostic and contain
no corpus-specific examples or seed rules.

\subsection{Initial Extraction Prompt}
\label{appx:extraction_prompt}

Prompt~1 is invoked once per regulatory section at Stage~2 of the pipeline. It
instructs the backbone LLM to decompose the source text into three typed components:
section metadata, definitions, and rule units, each conforming to a fixed schema. The
prompt encodes quality constraints inline, including null-filtering for non-actionable
sections and negation-preserving rules for action fields, to minimize the number of
judge-triggered repairs on well-formed inputs.

\begin{tcolorbox}[
  enhanced, breakable,
  colback=gray!3, colframe=red!55!black, boxrule=0.6pt, arc=4pt,
  left=6pt, right=6pt, top=0pt, bottom=5pt,
  attach boxed title to top left={yshift=-2.5mm, xshift=4mm},
  boxed title style={enhanced, colback=red!62!black, colframe=red!55!black,
    boxrule=0.5pt, arc=3pt, left=5pt, right=5pt, top=2pt, bottom=2pt},
  title={\textbf{\small\sffamily\color{white} Prompt 1
    \textnormal{\color{white!80!red}$\mid$} Rule Generation (Stage~2)}},
  label=prompt:generate,
  fontupper=\fontsize{6.2}{8}\selectfont\sffamily,
]
\bigskip
You are an expert at analyzing regulatory documents to extract structured, actionable
rules from \{\}.

\medskip
\noindent\textbf{\textcolor{red!65!black}{\#\# Task}}\\
Extract all rules, requirements, obligations, prohibitions, permissions, and procedures.
Each distinct rule = one RuleUnit.

\smallskip
\noindent\textbf{Requirements:}\\
\textbullet\ \textbf{Exhaustive}: Capture every rule in the text\\
\textbullet\ \textbf{Objective}: Use only source text -- no inference or interpretation\\
\textbullet\ \textbf{Precise}: Include verbatim quotes to ground extractions\\
\textbullet\ \textbf{Structured}: Decompose into clear components

\medskip
\noindent\textbf{\textcolor{red!65!black}{\#\# Extraction Guidelines}}

\smallskip
\noindent\textbf{\textcolor{gray!55!black}{\#\#\# Section Metadata}}

\smallskip
\noindent\textbf{Citation}\\
\textbullet\ Official regulatory section number (e.g., "17 CFR 275.0-2", "Rule 144",
"\S230.405")\\
\textbullet\ DO NOT use Federal Register citations (e.g., "51 FR 32907") as
\texttt{section\_cite}\\
\textbullet\ If no regulatory section identifier exists in source, set to null

\smallskip
\noindent\textbf{Title}\\
\textbullet\ Section's official heading

\smallskip
\noindent\textbf{Effective Dates}\\
\textbullet\ Extract dates in EXACT format from source (e.g., "Sept. 17, 1986" NOT
"September 17, 1986")\\
\textbullet\ Preserve abbreviations, punctuation, and spacing\\
\textbullet\ Extract ALL dates with appropriate event types\\
\textbullet\ Multiple FR citations typically indicate: first = "adopted", subsequent =
"amended"\\
\textbullet\ Look for context clues in brackets (e.g., "amended at 64 FR...")\\
\textbullet\ Common event types: adopted, amended, effective, rescinded

\smallskip
\noindent\textbf{notes}\\
\textbullet\ Set to null unless source contains additional context not captured in other
fields

\smallskip
\noindent\textbf{x\_extensions}\\
\textbullet\ Set to null unless source contains non-standard metadata

\smallskip
\noindent\textbf{Definitions}\\
Extract terms whose meaning is established by the text:\\
\textbullet\ Positive definitions: "X means Y"\\
\textbullet\ Negative definitions: "X is not Y" or "X does not include Y"\\
\textbullet\ Example: "A transaction...is not an assignment" $\rightarrow$ defines what
"assignment" excludes

\smallskip
\noindent Note: Definitional text may ALSO constitute a rule (e.g., "X is not
considered Y" both defines Y's scope AND exempts X). Extract to BOTH
\texttt{definitions} and \texttt{extracted\_rules} when applicable.

\medskip
\noindent\textbf{\textcolor{gray!55!black}{\#\#\# For Each Rule Extract:}}

\smallskip
\noindent\textbf{\#\#\#\# 1. Rule Identification}\\
\textbullet\ \textbf{rule\_id}: Unique identifier (e.g., "RULE-001"). Use consistent
nomenclature across document.\\
\textbullet\ \textbf{label}: Concise summary (5--25 words)\\
\textbullet\ \textbf{rule\_type}: obligation | prohibition | permission | exemption |
definition-application | safe-harbor | procedure | clarification | deeming |
condition-precedent | other (provide \texttt{other\_label})\\
\hspace{4mm}\textbullet\ Use \textbf{exemption} for: Carves out exceptions or states
what is NOT covered (e.g., "X is not deemed Y")\\
\hspace{4mm}\textbullet\ Use \textbf{clarification} for: Explains scope or meaning
without imposing new obligations

\medskip
\noindent\textbf{\#\#\#\# 2. Targets (WHO must comply)}\\
\textbf{CRITICAL: Target = WHO is being instructed to perform the action per this rule,
NOT who is the action being performed on!}

\smallskip
\noindent Example:\\
\textbullet\ "Issuers must file to Commission" $\rightarrow$ Target is \textbf{issuer}
(not Commission)\\
\textbullet\ "Any person may serve by furnishing Commission..." $\rightarrow$ Target is
\textbf{any\_person}\\
\textbullet\ "If $\langle$event$\rangle$, the head of Commission must..." $\rightarrow$
Target is \textbf{head of Commission} (not Commission)

\smallskip
\noindent\textbf{For each target:}\\
\textbullet\ \textbf{role}: answers the "who must follow this rule?" question\\
\textbullet\ \textbf{Preserve exact entity names}: "Secretary of the Commission" NOT
"commission"\\
\textbullet\ \textbf{qualifiers}: captures any additional conditions that apply
(e.g. "with 15+ employees")

\smallskip
\noindent\textbf{IMPORTANT -- Resolve Vague References:}\\
\textbullet\ If source text uses "you", "the registrant", "such person", determine the
actual entity from:\\
\hspace{4mm}-- Section title (e.g., "Rules for Investment Advisers" $\rightarrow$
target is "investment adviser")\\
\hspace{4mm}-- Regulatory context (e.g., Form CRS is for broker-dealers and investment
advisers)\\
\hspace{4mm}-- Earlier definitional text in the section

\medskip
\noindent\textbf{\#\#\#\# 3. Statement}\\
Extract the complete rule decomposition as a single structured object with the
following fields:

\smallskip
\noindent\textbf{3.1 action}\\
\textbullet\ Primary regulatory verb phrase (multi-word OK: "file reports", "serve
legal process")\\
\textbullet\ \textbf{CRITICAL: For negative statements, INCLUDE the negation in the
action field}\\
\textbullet\ WHAT must/must not/may be done?

\smallskip
\noindent\textbf{Examples of negative actions:}
"is not deemed"\ \ \textbullet\ "does not constitute"\ \
\textbullet\ "shall not be required"\ \ \textbullet\ "are not subject to"

\smallskip
\noindent\textcolor{red!70!black}{$\times$}\ Wrong: "deemed"
\qquad\textcolor{green!50!black}{$\checkmark$}\ Correct: "is not deemed"\\
\textcolor{red!70!black}{$\times$}\ Wrong: "considered"
\qquad\textcolor{green!50!black}{$\checkmark$}\ Correct: "is not considered"

\smallskip
\noindent\textbf{3.2 action\_object}\\
\textbullet\ WHO/WHAT is acted upon (e.g., "to the Commission", "non-resident advisers",
"assignment")

\smallskip
\noindent\textbf{3.3 method\_or\_conditions}\\
\textbullet\ HOW performed -- mechanisms, procedures (e.g., "by furnishing documents",
"by delivering the amended Form CRS or by communicating through another disclosure")\\
\textbullet\ \textbf{DO NOT include temporal/quantitative constraints here} -- extract
those to \texttt{constraints} field

\smallskip
\noindent\textbf{3.4 constraints}\\
Extract temporal/quantitative/qualitative limits as an array.
\textbf{If none, return [].}\\
For each constraint:\\
\textbullet\ \textbf{text}: Clear description (e.g., "within 10 business days", "not
less than \$5,000", "without charge")\\
\textbullet\ \textbf{applies\_to}: What is constrained (null if ambiguous)

\smallskip
\noindent\textbf{3.5 conditions}\\
Capture ALL conditions that define when, where, or under what circumstances the rule
applies. \textbf{If unconditional, return [].}

\smallskip
\noindent\textbf{IMPORTANT: The trigger should describe the QUALIFYING circumstance,
not just repeat the subject}\\
\textcolor{red!70!black}{$\times$}\ Wrong: trigger: "transaction"\\
\textcolor{green!50!black}{$\checkmark$}\ Correct: trigger: "transaction does not
result in change of actual control or management"

\smallskip
\noindent\textbf{Types of conditions to capture:}

\noindent 1.\ \textbf{Event-based triggers} (something happens):\\
\textbullet\ "If process is served..." $\rightarrow$ trigger: "process is served on the
Commission"\\
\textbullet\ "When trading volume exceeds..." $\rightarrow$ trigger: "trading volume
exceeds threshold"\\
\textbullet\ "Upon receipt of notice..." $\rightarrow$ trigger: "receipt of notice"

\noindent 2.\ \textbf{Scope/jurisdictional conditions} (regulatory framework):\\
\textbullet\ "Under Forms ADV and ADV-NR..." $\rightarrow$ trigger: "Under Forms ADV
and ADV-NR"\\
\textbullet\ "For purposes of Section 5..." $\rightarrow$ trigger: "For purposes of
Section 5"\\
\textbullet\ "As provided in Regulation S-K..." $\rightarrow$ trigger: "As provided
in Regulation S-K"

\noindent 3.\ \textbf{Status-based conditions} (entity characteristics):\\
\textbullet\ "If the issuer is foreign..." $\rightarrow$ trigger: "issuer is foreign"\\
\textbullet\ "For reporting companies..." $\rightarrow$ trigger: "entity is a reporting
company"

\smallskip
\noindent For each condition:\\
\textbullet\ \textbf{trigger}: The condition that must be met (can be event, scope, or
status)\\
\textbullet\ \textbf{time\_window}: Start/end times and timezone (only for temporal
conditions)\\
\textbullet\ \textbf{cross\_references}: Regulatory citations mentioned in the condition
(e.g., "section 205(a)(2) of the Act")

\smallskip
\noindent\textbf{3.6 exceptions}\\
Extract exemptions/carve-outs. \textbf{If none, return [].}\\
For each exception:\\
\textbullet\ \textbf{text}: Exception description\\
\textbullet\ \textbf{cross\_references}: Regulatory citations elaborating exception

\smallskip
\noindent\textbf{3.7 penalties\_or\_consequences}\\
Extract stated consequences of non-compliance. \textbf{If none, return null.}\\
For each penalty/consequence:\\
\textbullet\ \textbf{text}: Description of penalty or consequence\\
\textbullet\ \textbf{cross\_references}: Regulatory citations related to the penalty

\smallskip
\noindent\textbf{3.8 purpose}\\
\textbullet\ Stated objective (only if explicit; else null)

\smallskip
\noindent\textbf{3.9 verbatim}\\
\textbullet\ \textbf{REQUIRED} -- Exact source quote establishing rule\\
\textbullet\ Must include ALL sentences that contribute to this RuleUnit

\medskip
\noindent\textbf{\#\#\#\# 4. Additional Elements}\\
\textbullet\ \textbf{citations}: ALL regulatory references mentioned in the rule,
regardless of where they appear in the statement. Include references from conditions,
exceptions, and the main statement. For each citation provide the reference text and
context of its use.\\
\textbullet\ \textbf{examples}: Illustrative examples from source (lift verbatim; else
null)

\medskip
\noindent\textbf{\textcolor{red!65!black}{\#\# Output Format}}\\
JSON conforming to \texttt{SecActSectionExtraction} schema:\\
\textbullet\ \texttt{section\_meta}: Citation, title, effective dates\\
\textbullet\ \texttt{definitions}: Terms defined in section with explanations\\
\textbullet\ \texttt{extracted\_rules}: List of RuleUnit objects

\medskip
\noindent\textbf{\textcolor{red!65!black}{\#\# Quality Standards}}\\
\textbullet\ \textbf{Completeness}: Every regulatory statement $\rightarrow$ RuleUnit
with all applicable fields filled\\
\textbullet\ \textbf{Fidelity}: Verbatim = actual quotes, not paraphrases\\
\textbullet\ \textbf{Granularity}: Multiple distinct obligations $\rightarrow$ separate
RuleUnits\\
\textbullet\ \textbf{No Hallucination}: Only explicit content; use null/[] when absent\\
\textbullet\ \textbf{Preserve Specificity}: Exact entity names, titles, "if" statements
-- no simplification\\
\textbullet\ \textbf{Regulatory Citations}: Federal Register citations (e.g., [51 FR
32907, Sept. 17, 1986]) indicate effective dates but should not be extracted to notes
or x\_extensions unless they provide substantive context beyond dating.

\medskip
\noindent\textbf{\textcolor{red!65!black}{\#\# Special Cases}}\\
\textbullet\ \textbf{Compound Rules}: Multiple obligations on different targets
$\rightarrow$ split into RuleUnits\\
\textbullet\ \textbf{Cross-References}: "as defined in \S240.10b-5" $\rightarrow$
capture in \texttt{cross\_references}\\
\textbullet\ \textbf{Implicit Requirements}: Extract only if clearly stated (e.g.,
"failure to file within 10 days requires..." implies filing obligation)

\medskip
\noindent\textbf{\textcolor{red!65!black}{\#\# CRITICAL OUTPUT REQUIREMENTS}}

\smallskip
\noindent Your JSON output MUST include these fields for EVERY RuleUnit
(use \texttt{[]} for empty):

\begin{tcolorbox}[
  colback=gray!13, colframe=gray!35, boxrule=0.4pt, arc=2pt,
  left=5pt, right=5pt, top=3pt, bottom=3pt,
  fontupper=\fontsize{5.8}{7.5}\selectfont\ttfamily,
]
\{\\
\hspace{3mm}"rule\_id": "...",\\
\hspace{3mm}"label": "...",\\
\hspace{3mm}"rule\_type": \{...\},\\
\hspace{3mm}"targets": [...],\\
\hspace{3mm}"statement": \{\\
\hspace{6mm}"action": "...",\\
\hspace{6mm}"action\_object": "...",\\
\hspace{6mm}"method": "...",\\
\hspace{6mm}"constraints": [...],\hspace{3mm}\textit{// REQUIRED -- use [] only if genuinely absent}\\
\hspace{6mm}"conditions": [...],\hspace{4mm}\textit{// REQUIRED -- use [] only if genuinely absent}\\
\hspace{6mm}"exceptions": [...],\hspace{4mm}\textit{// REQUIRED -- use [] only if genuinely absent}\\
\hspace{6mm}"penalties\_or\_consequences": [...],\hspace{3mm}\textit{// REQUIRED -- use null only if genuinely absent}\\
\hspace{6mm}"purpose": "...",\\
\hspace{6mm}"verbatim": "..."\\
\hspace{3mm}\},\\
\hspace{3mm}"citations": \{...\},\\
\hspace{3mm}"examples": [...]\\
\}
\end{tcolorbox}

\smallskip
\noindent Missing any of these fields will cause validation failure.

\medskip
\noindent\textbf{\textcolor{red!65!black}{FINAL VERIFICATION PROTOCOL}}

\smallskip
\noindent Before returning your JSON output:

\noindent\textbf{1. Section Metadata Verification:}\\
\textbullet\ \texttt{section\_cite}: null if not in source, exact format if present
(DO NOT use FR citations)\\
\textbullet\ \texttt{effective\_dates}: ALL dates with correct event types ("adopted"
for first, "amended" for subsequent)\\
\textbullet\ Dates: exact format match including abbreviations (e.g., "Sept." not
"September")\\
\textbullet\ definitions: Extract negative definitions too (e.g., "X is not Y")\\
\textbullet\ Optional fields (notes, x\_extensions): explicitly null if unused

\noindent\textbf{2. Rule Extraction Verification:}\\
\textbullet\ action: For negative statements, verify negation is INCLUDED (e.g., "is
not deemed" NOT "deemed")\\
\textbullet\ rule\_type: "exemption" for exclusionary rules (e.g., "X is not Y")\\
\textbullet\ conditions: trigger describes the qualifying circumstance, not just the
subject\\
\textbullet\ conditions: \texttt{cross\_references} populated if regulatory citations
mentioned\\
\textbullet\ citations: ALL regulatory references included with context

\noindent\textbf{3. Completeness Check:} For each RuleUnit, verify that you've
populated:\\
\textbullet\ conditions array (even if single item)\\
\textbullet\ constraints array (check for temporal/quantitative limits)\\
\textbullet\ exceptions array (check for carve-outs)\\
\textbullet\ penalties\_or\_consequences (check for stated consequences)

\noindent\textbf{4. Source Cross-Check:} For any field marked as \texttt{[]}, re-read
the verbatim quote and confirm no extractable information exists.

\noindent\textbf{5. Scope Conditions:} Specifically check if the rule begins with
regulatory context like "Under [Form/Regulation]" or "For purposes of [Section]" --
this is ALWAYS a condition.

\medskip
\noindent Now analyze the document and extract all rules:\\
\{\}
\end{tcolorbox}
\vspace{1mm}
{\small\sffamily\textit{Prompt~1: Initial rule generation prompt (Stage~2 of
De~Jure). The first placeholder \texttt{\{\}} is filled with the regulatory domain;
the second with the source section text. No domain-specific examples or seed rules
are included.}}

\subsection{Regeneration Prompts}
\label{appx:regen_prompts}

When a pipeline stage fails to meet the acceptance threshold $\theta$, the backbone
LLM is re-prompted with three inputs: the original source text, the failing
extraction, and the judge's structured critique. Three stage-specific regeneration
prompts are used, one per judgment stage, each targeting the exact output type of its
corresponding judge. Crucially, each prompt asks the model to correct only the
deficient fields identified in the critique rather than regenerating the full
extraction, keeping repairs surgical and computationally bounded.

\begin{tcolorbox}[
  enhanced, breakable,
  colback=gray!3, colframe=red!55!black, boxrule=0.6pt, arc=4pt,
  left=6pt, right=6pt, top=0pt, bottom=5pt,
  attach boxed title to top left={yshift=-2.5mm, xshift=4mm},
  boxed title style={enhanced, colback=red!62!black, colframe=red!55!black,
    boxrule=0.5pt, arc=3pt, left=5pt, right=5pt, top=2pt, bottom=2pt},
  title={\textbf{\small\sffamily\color{white} Prompt 2
    \textnormal{\color{white!80!red}$\mid$} Stage~1 Regeneration: Section Metadata}},
  fontupper=\fontsize{6.2}{8}\selectfont\sffamily,
]
\bigskip
Your task is to extract section metadata from a source text containing regulatory
content. You have been provided with a previous extraction that was assessed based on
specific criteria and determined to be incorrect.

\medskip
\noindent Using the critique, generate corrected metadata from the source text to
achieve the highest score on all mentioned criteria:\\
\textbullet\ Completeness\\
\textbullet\ Fidelity to Source Text\\
\textbullet\ Non-Hallucination\\
\textbullet\ Title Quality\\
\textbullet\ Precision of Citations and Dates\\
\textbullet\ Reasonable Population of Optional Fields

\medskip
\noindent\textbf{\textcolor{red!65!black}{Follow the format below:}}
\begin{tcolorbox}[colback=gray!13, colframe=gray!35, boxrule=0.4pt, arc=2pt,
  left=5pt, right=5pt, top=3pt, bottom=3pt,
  fontupper=\fontsize{5.8}{7.5}\selectfont\ttfamily]
\{\\
\hspace{3mm}"section\_cite": "citation from source",\\
\hspace{3mm}"title": "section title",\\
\hspace{3mm}"effective\_dates": [\{"event": "event\_type", "date": "exact date format"\}],\\
\hspace{3mm}"notes": "additional context if present in source",\hspace{2mm}\textit{// (optional)}\\
\hspace{3mm}"x\_extensions": \{\}\hspace{2mm}\textit{// (optional)}\\
\}
\end{tcolorbox}

\smallskip
\noindent Only generate the corrected metadata JSON and nothing else.

\medskip
\noindent\textbf{\textcolor{red!65!black}{Inputs:}}\\
Source Text: \{\}\\
Incorrect Metadata: \{\}\\
Critique: \{\}
\end{tcolorbox}
\vspace{1mm}
{\small\sffamily\textit{Prompt~2: Stage~1 regeneration prompt. Placeholders are filled
at runtime with the source section text, the failing metadata extraction, and the
Judge~1 critique.}}

\bigskip

\begin{tcolorbox}[
  enhanced, breakable,
  colback=gray!3, colframe=red!55!black, boxrule=0.6pt, arc=4pt,
  left=6pt, right=6pt, top=0pt, bottom=5pt,
  attach boxed title to top left={yshift=-2.5mm, xshift=4mm},
  boxed title style={enhanced, colback=red!62!black, colframe=red!55!black,
    boxrule=0.5pt, arc=3pt, left=5pt, right=5pt, top=2pt, bottom=2pt},
  title={\textbf{\small\sffamily\color{white} Prompt 3
    \textnormal{\color{white!80!red}$\mid$} Stage~2 Regeneration: Definitions}},
  fontupper=\fontsize{6.2}{8}\selectfont\sffamily,
]
\bigskip
Your task is to extract definitions from a source text containing regulatory content.
You have been provided with a previous extraction that was assessed based on specific
criteria and determined to be incorrect.

\medskip
\noindent Using the critique, generate corrected definitions from the source text to
achieve the highest score on all mentioned criteria:\\
\textbullet\ Completeness\\
\textbullet\ Fidelity to Source Text\\
\textbullet\ No Hallucination or Fabrication\\
\textbullet\ Precision and Formatting\\
\textbullet\ Quality of Terms

\medskip
\noindent\textbf{\textcolor{red!65!black}{Follow the format below:}}
\begin{tcolorbox}[colback=gray!13, colframe=gray!35, boxrule=0.4pt, arc=2pt,
  left=5pt, right=5pt, top=3pt, bottom=3pt,
  fontupper=\fontsize{5.8}{7.5}\selectfont\ttfamily]
\{\\
\hspace{3mm}"definitions": [\\
\hspace{6mm}\{\\
\hspace{9mm}"term": term,\\
\hspace{9mm}"text": definition\\
\hspace{6mm}\}\\
\hspace{3mm}]\\
\}
\end{tcolorbox}

\smallskip
\noindent Only generate the corrected definitions JSON and nothing else.

\medskip
\noindent\textbf{\textcolor{red!65!black}{Inputs:}}\\
Source Text: \{\}\\
Incorrect Definitions: \{\}\\
Critique: \{\}
\end{tcolorbox}
\vspace{1mm}
{\small\sffamily\textit{Prompt~3: Stage~2 regeneration prompt. Placeholders are filled
at runtime with the source section text, the failing definitions extraction, and the
Judge~2 critique.}}

\bigskip

\begin{tcolorbox}[
  enhanced, breakable,
  colback=gray!3, colframe=red!55!black, boxrule=0.6pt, arc=4pt,
  left=6pt, right=6pt, top=0pt, bottom=5pt,
  attach boxed title to top left={yshift=-2.5mm, xshift=4mm},
  boxed title style={enhanced, colback=red!62!black, colframe=red!55!black,
    boxrule=0.5pt, arc=3pt, left=5pt, right=5pt, top=2pt, bottom=2pt},
  title={\textbf{\small\sffamily\color{white} Prompt 4
    \textnormal{\color{white!80!red}$\mid$} Stage~3 Regeneration: Rule Units}},
  fontupper=\fontsize{6.2}{8}\selectfont\sffamily,
]
\bigskip
Your task is to extract rule units from a source text containing regulatory content.
You have been provided with a previous extraction that was assessed based on specific
criteria and determined to be incorrect.

\medskip
\noindent Using the critique, generate a corrected rule unit from the source text to
achieve the highest score on all mentioned criteria:\\
\textbullet\ Completeness\\
\textbullet\ Conciseness (for label)\\
\textbullet\ Accuracy (of rule\_type)\\
\textbullet\ Consistency (of targets)\\
\textbullet\ Fidelity to Source Text (statements)\\
\textbullet\ Neutrality\\
\textbullet\ Actionability\\
\textbullet\ No Hallucination

\medskip
\noindent\textbf{\textcolor{red!65!black}{Follow the format below:}}
\begin{tcolorbox}[colback=gray!13, colframe=gray!35, boxrule=0.4pt, arc=2pt,
  left=5pt, right=5pt, top=3pt, bottom=3pt,
  fontupper=\fontsize{5.8}{7.5}\selectfont\ttfamily]
\{\\
\hspace{3mm}"rule\_id": "rule\_id from the source/generated",\\
\hspace{3mm}"label": "concise summary (5--25 words)",\\
\hspace{3mm}"rule\_type": "obligation" | "prohibition" | "permission" | "exemption" |\\
\hspace{14mm}"definition-application" | "safe-harbor" | "procedure" |\\
\hspace{14mm}"clarification" | "deeming" | "condition-precedent" | "other",\\
\hspace{3mm}"targets": ["WHO must comply, is prohibited, or is granted permission"],\\
\hspace{3mm}"statement": \{\\
\hspace{6mm}"action": "primary regulatory action",\\
\hspace{6mm}"action\_object": "direct object or recipient of the action",\\
\hspace{6mm}"method": "HOW the action must be performed",\\
\hspace{6mm}"constraints": [...],\hspace{3mm}\textit{// REQUIRED -- [] if genuinely absent}\\
\hspace{6mm}"conditions": [...],\hspace{4mm}\textit{// REQUIRED -- [] if genuinely absent}\\
\hspace{6mm}"exceptions": [...],\hspace{4mm}\textit{// REQUIRED -- [] if genuinely absent}\\
\hspace{6mm}"penalties\_or\_consequences": [...],\hspace{3mm}\textit{// REQUIRED -- null if genuinely absent}\\
\hspace{6mm}"purpose": "stated objective",\\
\hspace{6mm}"verbatim": "source quote establishing rule"\\
\hspace{3mm}\},\\
\hspace{3mm}"citations": [...],\hspace{3mm}\textit{// REQUIRED -- [] if genuinely absent}\\
\hspace{3mm}"examples": [...]\hspace{4mm}\textit{// REQUIRED -- [] if genuinely absent}\\
\}
\end{tcolorbox}

\smallskip
\noindent Only generate the corrected RuleUnit JSON and nothing else.

\medskip
\noindent\textbf{\textcolor{red!65!black}{Inputs:}}\\
Source Text: \{\}\\
Incorrect RuleUnit: \{\}\\
Critique: \{\}
\end{tcolorbox}
\vspace{1mm}
{\small\sffamily\textit{Prompt~4: Stage~3 regeneration prompt. Placeholders are filled
at runtime with the source section text, the failing rule unit extraction, and the
Judge~3 critique.}}

\subsection{Judge Prompts}
\label{appx:judge_prompts}

The three judge prompts below are invoked sequentially during the multi-criteria
evaluation stage to assess section metadata (Judge~1), definitions (Judge~2), and
individual rule units (Judge~3). Each prompt instructs the judge LLM to score the
extraction independently on per-stage criteria and to produce structured
natural-language critiques. These critiques are subsequently passed verbatim as input
to the corresponding regeneration prompts in Section~\ref{appx:regen_prompts},
forming the closed judge-repair loop at the core of De~Jure. All prompts share the
same 0--5 scoring rubric and instruct the judge to assign 5 to inapplicable criteria,
ensuring the threshold $\theta$ is not penalized by structural variation across
regulatory sections.

\begin{tcolorbox}[
  enhanced, breakable,
  colback=gray!3, colframe=red!55!black, boxrule=0.6pt, arc=4pt,
  left=6pt, right=6pt, top=0pt, bottom=5pt,
  attach boxed title to top left={yshift=-2.5mm, xshift=4mm},
  boxed title style={enhanced, colback=red!62!black, colframe=red!55!black,
    boxrule=0.5pt, arc=3pt, left=5pt, right=5pt, top=2pt, bottom=2pt},
  title={\textbf{\small\sffamily\color{white} Prompt 5
    \textnormal{\color{white!80!red}$\mid$} Judge~1: Section Metadata Evaluation}},
  fontupper=\fontsize{6.2}{8}\selectfont\sffamily,
]
\bigskip
Evaluate whether the extracted section metadata is substantially accurate based on
the following criteria:

\medskip
\noindent\textbf{\textcolor{red!65!black}{1. Completeness}}\\
Major metadata elements should be extracted and populated:\\
\textbullet\ \texttt{section\_cite} should be present and identify the correct section\\
\textbullet\ \texttt{title} should be captured if clearly present in the source\\
\textbullet\ \texttt{effective\_dates} should include at least the primary temporal event
(if any)\\
\textbullet\ \texttt{notes} (optional) may capture additional context\\
\textbullet\ \texttt{x\_extensions} (optional) may include non-standard metadata\\
Missing critical fields (\texttt{section\_cite}, \texttt{title} when present) are
significant issues. Missing optional fields or secondary dates are minor issues.

\smallskip
\noindent\textbf{\textcolor{red!65!black}{2. Fidelity to Source Text}}\\
Notes and x\_extensions should reasonably reflect the source content. Direct quotes,
close paraphrasing, or reasonable interpretations are all acceptable. Minor rewording
or normalization of language is acceptable.

\smallskip
\noindent\textbf{\textcolor{red!65!black}{3. Non-Hallucination}}\\
Fields should only be populated when corresponding information exists in the source.
Do not fabricate dates, citations, or contextual information. Event types in
\texttt{effective\_dates} should be grounded in the source (normalized terminology
acceptable, e.g., "enacted" $\rightarrow$ "adopted"). This criterion is strict:
hallucinated information is a significant problem.

\smallskip
\noindent\textbf{\textcolor{red!65!black}{4. Title Quality}}\\
Title should accurately reflect the section title if present. Minor formatting
variations are acceptable. Null is appropriate if no title exists.

\smallskip
\noindent\textbf{\textcolor{red!65!black}{5. Precision of Citations and Dates}}\\
Section citations should identify the correct section (minor formatting differences
acceptable). Dates should be correct to at least the month/year level. If multiple
effective dates exist, capturing the primary date is essential; missing secondary
dates is a minor issue.

\smallskip
\noindent\textbf{\textcolor{red!65!black}{6. Reasonable Population of Optional Fields}}\\
When \texttt{notes} or \texttt{x\_extensions} are populated, they should add value.
Omitting these fields when relevant information exists is a minor issue, not a major
deficiency.

\medskip
\noindent\textbf{\textcolor{red!65!black}{Scoring Guidelines (per criterion):}}\\
\textbullet\ \textbf{5.0}: Fully satisfied with no errors\\
\textbullet\ \textbf{4.0--4.9}: Mostly satisfied with minor issues\\
\textbullet\ \textbf{3.0--3.9}: Partially satisfied with notable gaps or minor
inaccuracies\\
\textbullet\ \textbf{2.0--2.9}: Poorly satisfied with significant omissions or errors\\
\textbullet\ \textbf{1.0--1.9}: Barely satisfied with major problems\\
\textbullet\ \textbf{0.0--0.9}: Not satisfied -- critical failures or fabrications\\
Score each of the 6 criteria independently. If a criterion is not applicable, assign
5. Report the average as the final score.

\medskip
\noindent\textbf{\textcolor{red!65!black}{Inputs:}}\\
Source Text: \{\}\\
Extracted Metadata: \{\}
\end{tcolorbox}
\vspace{1mm}
{\small\sffamily\textit{Prompt~5: Judge~1 evaluation prompt for section metadata
(6 criteria). Output scores and critiques are consumed by Prompt~2.}}

\bigskip

\begin{tcolorbox}[
  enhanced, breakable,
  colback=gray!3, colframe=red!55!black, boxrule=0.6pt, arc=4pt,
  left=6pt, right=6pt, top=0pt, bottom=5pt,
  attach boxed title to top left={yshift=-2.5mm, xshift=4mm},
  boxed title style={enhanced, colback=red!62!black, colframe=red!55!black,
    boxrule=0.5pt, arc=3pt, left=5pt, right=5pt, top=2pt, bottom=2pt},
  title={\textbf{\small\sffamily\color{white} Prompt 6
    \textnormal{\color{white!80!red}$\mid$} Judge~2: Definitions Evaluation}},
  fontupper=\fontsize{6.2}{8}\selectfont\sffamily,
]
\bigskip
Evaluate whether the extracted definitions and related rule extractions are
substantially accurate based on the following criteria:

\medskip
\noindent\textbf{\textcolor{red!65!black}{1. Completeness}}\\
Major definitional statements should be captured, including:\\
\textbullet\ Primary positive definitions ("X means Y", "X includes Y")\\
\textbullet\ Significant negative definitions or exclusions ("X is not Y", "X does
not include Y")\\
\textbullet\ Important context such as major exceptions or limitations\\
Missing primary definitions or major exclusions are significant issues. Missing minor
qualifications or secondary cross-references are minor issues.

\smallskip
\noindent\textbf{\textcolor{red!65!black}{2. Fidelity to Source Text}}\\
Extracted terms and definitions should reasonably reflect the source. Direct quotes,
close paraphrasing, or reasonable interpretations preserving core meaning are all
acceptable. Definitions should not substantially contradict the source or
misrepresent the term's meaning.

\smallskip
\noindent\textbf{\textcolor{red!65!black}{3. No Hallucination or Fabrication}}\\
Extract only definitions present in the source. Do not invent terms, definitions, or
context not grounded in the text. This criterion is strict: fabricated content is a
significant problem. Reasonable interpretations of existing content are acceptable.

\smallskip
\noindent\textbf{\textcolor{red!65!black}{4. Precision and Formatting}}\\
Terms should be substantially accurate in spelling and punctuation. Important
references (e.g., to statutes or sections) should be captured. Each extracted
definition should be clear and understandable. Minor formatting inconsistencies are
acceptable if core meaning is preserved.

\smallskip
\noindent\textbf{\textcolor{red!65!black}{5. Quality of Terms}}\\
Each extracted term should reasonably match the terminology used in the source.
Terms should accurately represent the intended meaning and context. Minor variations
in term format or phrasing are acceptable if they do not misrepresent the definition.

\medskip
\noindent\textbf{\textcolor{red!65!black}{Scoring Guidelines (per criterion):}}\\
\textbullet\ \textbf{5.0}: Fully satisfied with no errors\\
\textbullet\ \textbf{4.0--4.9}: Mostly satisfied with minor issues\\
\textbullet\ \textbf{3.0--3.9}: Partially satisfied with notable gaps or minor
inaccuracies\\
\textbullet\ \textbf{2.0--2.9}: Poorly satisfied with significant omissions or errors\\
\textbullet\ \textbf{1.0--1.9}: Barely satisfied with major problems\\
\textbullet\ \textbf{0.0--0.9}: Not satisfied -- critical failures or fabrications\\
Score each of the 5 criteria independently. If a criterion is not applicable, assign
5. Report the average as the final score.

\medskip
\noindent\textbf{\textcolor{red!65!black}{Inputs:}}\\
Source Text: \{\}\\
Extracted Definitions: \{\}
\end{tcolorbox}
\vspace{1mm}
{\small\sffamily\textit{Prompt~6: Judge~2 evaluation prompt for definitions
(5 criteria). Output scores and critiques are consumed by Prompt~3.}}

\bigskip

\begin{tcolorbox}[
  enhanced, breakable,
  colback=gray!3, colframe=red!55!black, boxrule=0.6pt, arc=4pt,
  left=6pt, right=6pt, top=0pt, bottom=5pt,
  attach boxed title to top left={yshift=-2.5mm, xshift=4mm},
  boxed title style={enhanced, colback=red!62!black, colframe=red!55!black,
    boxrule=0.5pt, arc=3pt, left=5pt, right=5pt, top=2pt, bottom=2pt},
  title={\textbf{\small\sffamily\color{white} Prompt 7
    \textnormal{\color{white!80!red}$\mid$} Judge~3: Rule Unit Evaluation}},
  fontupper=\fontsize{6.2}{8}\selectfont\sffamily,
]
\bigskip
You are an expert evaluator assessing structured rule extraction from regulatory
documents. You will receive the original source text and an extracted RuleUnit.
Evaluate each component based on the following criteria:

\medskip
\noindent\textbf{\textcolor{red!65!black}{1. Completeness}}\\
Core components (\texttt{label}, \texttt{rule\_type}, \texttt{targets},
\texttt{action}, \texttt{action\_object}) are significant if missing. Secondary
components (\texttt{method}, \texttt{constraints}, \texttt{conditions}) and optional
fields when absent are minor issues.

\smallskip
\noindent\textbf{\textcolor{red!65!black}{2. Conciseness}}\\
The label should be reasonably brief, summarizing the rule while preserving important
meaning. Slight wordiness is acceptable. Significant deviation from the rule's meaning
or scope should be noted.

\smallskip
\noindent\textbf{\textcolor{red!65!black}{3. Accuracy}}\\
The \texttt{rule\_type} should reasonably represent the type of rule in the source.
Minor classification judgment calls are acceptable. Significant misclassification that
misrepresents the rule's fundamental nature is a problem.

\smallskip
\noindent\textbf{\textcolor{red!65!black}{4. Consistency}}\\
Targets should reasonably align with the source. Minor terminology variations that
preserve meaning are acceptable. Targets should not significantly contradict or
misrepresent who the rule applies to.

\smallskip
\noindent\textbf{\textcolor{red!65!black}{5. Fidelity to Source Text}}\\
Statement components should reasonably reflect the source. Conditions, exceptions, and
constraints should capture the primary requirements. Minor deviations in phrasing that
do not affect legal interpretation are acceptable. Significant alterations of scope or
omission of critical qualifiers should be noted.

\smallskip
\noindent\textbf{\textcolor{red!65!black}{6. Neutrality}}\\
Labels and statements should present the source in a balanced manner. Minor
interpretive choices are acceptable. Significant bias or misrepresentation of the
rule's intent should be noted.

\smallskip
\noindent\textbf{\textcolor{red!65!black}{7. Actionability}}\\
Action and action\_object should provide reasonably clear guidance, understandable and
usable in a business context. Minor ambiguity is acceptable if the core action and
object are identifiable. Excessive abstraction that obscures what must be done is
problematic.

\smallskip
\noindent\textbf{\textcolor{red!65!black}{8. Non-Hallucination}}\\
Extract only rule units present in the source. Do not invent rule components,
conditions, targets, or context not grounded in the text. This criterion is strict:
fabricated content is a significant problem. Reasonable interpretations of existing
content are acceptable.

\medskip
\noindent\textbf{\textcolor{red!65!black}{Scoring Guidelines (per criterion):}}\\
\textbullet\ \textbf{5.0}: Fully satisfied with no errors\\
\textbullet\ \textbf{4.0--4.9}: Mostly satisfied with minor issues\\
\textbullet\ \textbf{3.0--3.9}: Partially satisfied with notable gaps or minor
inaccuracies\\
\textbullet\ \textbf{2.0--2.9}: Poorly satisfied with significant omissions or errors\\
\textbullet\ \textbf{1.0--1.9}: Barely satisfied with major problems\\
\textbullet\ \textbf{0.0--0.9}: Not satisfied -- critical failures or fabrications\\
Score each of the 8 criteria independently. If a criterion is not applicable, assign
5. Report the average as the final score. Focus on significant errors that materially
affect accuracy, completeness, or usability of the extracted rule.

\medskip
\noindent\textbf{\textcolor{red!65!black}{Inputs:}}\\
Source Text: \{\}\\
Extracted RuleUnit: \{\}
\end{tcolorbox}
\vspace{1mm}
{\small\sffamily\textit{Prompt~7: Judge~3 evaluation prompt for per-rule quality
(8 criteria). Output scores and critiques are consumed by Prompt~4.}}

\subsection{Question Generation Prompt}
\label{appx:question_gen_prompt}

The following prompt is used to generate evaluation questions for Experiment~3
(Section~\ref{sec:downstream-eval}). Qwen3-VL-8B-Instruct is prompted with
 HIPAA regulatory document to produce compliance-grounded questions fully
answerable from the passage alone, spanning factual, conditional, and analytical
types. The question count and passage content are injected at the \texttt{\{\}}
placeholders.

\begin{tcolorbox}[
  enhanced, breakable,
  colback=gray!3, colframe=red!55!black, boxrule=0.6pt, arc=4pt,
  left=6pt, right=6pt, top=0pt, bottom=5pt,
  attach boxed title to top left={yshift=-2.5mm, xshift=4mm},
  boxed title style={enhanced, colback=red!62!black, colframe=red!55!black,
    boxrule=0.5pt, arc=3pt, left=5pt, right=5pt, top=2pt, bottom=2pt},
  title={\textbf{\small\sffamily\color{white} Prompt 8
    \textnormal{\color{white!80!red}$\mid$} Question Generation: Downstream RAG Evaluation}},
  fontupper=\fontsize{6.2}{8}\selectfont\sffamily,
]
\bigskip
You are a senior legal analyst and compliance expert with over two decades of
experience specializing in U.S.\ healthcare law, HIPAA regulations (45 CFR Parts 160
and 164), and regulatory compliance frameworks.

\medskip
\noindent\textbf{\textcolor{red!65!black}{Areas of Expertise:}}\\
\textbullet\ HIPAA Privacy Rule, Security Rule, and Breach Notification Rule\\
\textbullet\ Protected Health Information (PHI) handling, use, and disclosure
requirements\\
\textbullet\ Covered entities, business associates, and their obligations\\
\textbullet\ Administrative, physical, and technical safeguards\\
\textbullet\ Enforcement, penalties, and compliance procedures

\medskip
Your role is to generate precise, grounded, and legally meaningful questions from
regulatory passages. You think like both a compliance officer and a litigator:
you understand nuance, edge cases, and the practical implications of regulatory
language. Given the following regulatory passage, generate exactly \{\} questions
that a compliance officer, legal analyst, or auditor might ask.

\medskip
\noindent\textbf{\textcolor{red!65!black}{Strict Rules:}}\\
\textbullet\ Every question must be directly and fully answerable from the passage
alone\\
\textbullet\ Do NOT introduce concepts, entities, or scenarios not present in the
passage\\
\textbullet\ Do NOT ask questions requiring outside knowledge or inference beyond
the passage\\
\textbullet\ Questions must be diverse: cover who, what, when, how, and under what
conditions\\
\textbullet\ Questions must be specific and unambiguous\\
\textbullet\ Questions should vary in complexity: mix factual, conditional, and
analytical types\\
\textbullet\ Each question must stand alone as a complete, clear sentence

\medskip
\noindent\textbf{\textcolor{red!65!black}{Output Format:}}\\
Return ONLY a valid Python list of \{\} strings. No explanation, no preamble, no
commentary.

\smallskip
\noindent Example:\\
\texttt{["Question one?", "Question two?", ...]}

\medskip
\noindent\textbf{\textcolor{red!65!black}{Input:}}\\
Passage: \{\}
\end{tcolorbox}
\vspace{1mm}
{\small\sffamily\textit{Prompt~8: Question generation prompt for downstream evaluation on compliance QA via RAG (Experiment~3). Injected with the target passage and desired question
count; output is a Python list used to construct the 100-question evaluation set.}}

\newpage
\section{Extraction Schema}
\label{appx:schema}

\definecolor{schemebg}{HTML}{F5F7FA}
\definecolor{schemeframe}{HTML}{2B5F8A}
\definecolor{schemetitle}{HTML}{1E4870}
\definecolor{schemeaccent}{HTML}{2B5F8A}

De Jure structures LLM outputs using guided JSON generation, with Pydantic-defined schemas enforcing the extraction of discrete rule units from the source text. 
\tcbset{
  schemebox/.style={
    enhanced, breakable,
    colback=schemebg, colframe=schemeframe, boxrule=0.6pt, arc=4pt,
    left=6pt, right=6pt, top=0pt, bottom=5pt,
    attach boxed title to top left={yshift=-2.5mm, xshift=4mm},
    boxed title style={enhanced, colback=schemetitle, colframe=schemeframe,
      boxrule=0.5pt, arc=3pt, left=5pt, right=5pt, top=2pt, bottom=2pt},
    fontupper=\fontsize{6.2}{8}\selectfont\sffamily,
  }
}

\begin{tcolorbox}[
  schemebox,
  title={\textbf{\small\sffamily\color{white} Schema~1
    \textnormal{\color{white!80!schemetitle}$\mid$}
    SectionExtraction \textnormal{\color{white!70!schemetitle}(top-level)}}},
]
\bigskip
\noindent Top-level container returned by the extraction pipeline.
All sub-objects use \texttt{additionalProperties:~false}.

\medskip
\noindent\textbf{\textcolor{schemeaccent}{Root fields}}\\[2pt]
\begin{tabular}{@{}l l l p{5.2cm}@{}}
\texttt{section\_meta}      & \textit{SectionMeta}         & \textbf{req} & Metadata about the regulatory section \\
\texttt{definitions}        & \textit{DefinitionEntry[ ]}  & opt           & Defined terms extracted from the section \\
\texttt{extracted\_rules}   & \textit{RuleUnit[ ]}         & \textbf{req} & List of structured rule extractions \\
\end{tabular}

\medskip
\noindent\textbf{\textcolor{schemeaccent}{SectionMeta}}\\[2pt]
\begin{tabular}{@{}l l l p{5.2cm}@{}}
\texttt{schema\_version}  & \textit{string}            & opt (``1.0.0'')  & Version of the extraction schema \\
\texttt{section\_cite}    & \textit{string}            & \textbf{req}     & Citation of the section \\
\texttt{title}            & \textit{string}            & \textbf{req}     & Title of the section \\
\texttt{effective\_dates} & \textit{EffectiveDate[ ]}  & opt              & Timeline of adoptions, amendments, rescissions \\
\texttt{notes}            & \textit{string}            & opt              & Additional context \\
\texttt{x\_extensions}    & \textit{object}            & opt              & Custom extension namespace \\
\end{tabular}

\medskip
\noindent\textbf{\textcolor{schemeaccent}{EffectiveDate}}\\[2pt]
\begin{tabular}{@{}l l l p{5.2cm}@{}}
\texttt{event}        & \textit{enum}   & \textbf{req} & \texttt{adopted} $|$ \texttt{amended} $|$ \texttt{rescinded} $|$ \texttt{note} \\
\texttt{date}         & \textit{string} & \textbf{req} & ISO date, e.g.\ \texttt{2023-02-27} \\
\texttt{fr\_citation} & \textit{string} & opt           & Federal Register citation, if available \\
\end{tabular}

\medskip
\noindent\textbf{\textcolor{schemeaccent}{DefinitionEntry}}\\[2pt]
\begin{tabular}{@{}l l l p{5.2cm}@{}}
\texttt{term}              & \textit{string}       & \textbf{req} & The term being defined \\
\texttt{text}              & \textit{string}       & \textbf{req} & Definition text from source \\
\texttt{scope}             & \textit{enum}         & opt           & \texttt{section} $|$ \texttt{part} $|$ \texttt{act} $|$ \texttt{context-dependent}; leave null if unstated \\
\texttt{cross\_references} & \textit{CrossRef[ ]}  & opt           & Related regulatory references \\
\end{tabular}
\end{tcolorbox}
\vspace{1mm}
{\small\sffamily\textit{Schema~1: Top-level extraction container with section
metadata and definitions. All objects enforce strict schemas
(\texttt{additionalProperties:~false}). Required ('req') and Optional ('opt') categories are tagged separately.}}

\bigskip

\begin{tcolorbox}[
  schemebox,
  title={\textbf{\small\sffamily\color{white} Schema~2
    \textnormal{\color{white!80!schemetitle}$\mid$}
    RuleUnit}},
]
\bigskip
\noindent A single extracted regulatory rule with type classification, targets,
decomposed statement, citations, and judge scores populated post-evaluation.

\medskip
\noindent\textbf{\textcolor{schemeaccent}{RuleUnit}}\\[2pt]
\begin{tabular}{@{}l l l p{5.2cm}@{}}
\texttt{rule\_id}      & \textit{string}         & \textbf{req} & Unique identifier for this rule \\
\texttt{label}         & \textit{string}         & \textbf{req} & Short summary of the rule \\
\texttt{rule\_type}    & \textit{RuleType}       & \textbf{req} & Classification of the rule \\
\texttt{targets}       & \textit{Target[ ]}      & \textbf{req} & Entities subject to the rule \\
\texttt{statement}     & \textit{Statement}      & \textbf{req} & Decomposed regulatory requirement \\
\texttt{citations}     & \textit{Citations}      & opt           & Supporting regulatory citations \\
\texttt{judge\_score}  & \textit{JudgeScore}     & opt           & Aggregated evaluation scores \\
\texttt{examples}      & \textit{string[ ]}      & opt           & Examples lifted from source text, if available \\
\end{tabular}

\medskip
\noindent\textbf{\textcolor{schemeaccent}{RuleType}}\\[2pt]
\begin{tabular}{@{}l l l p{5.2cm}@{}}
\texttt{type}         & \textit{enum} & \textbf{req} &
  \texttt{obligation} $|$ \texttt{prohibition} $|$ \texttt{permission} $|$
  \texttt{exemption} $|$ \texttt{definition-application} $|$
  \texttt{safe-harbor} $|$ \texttt{procedure} $|$ \texttt{clarification} $|$
  \texttt{deeming} $|$ \texttt{condition-precedent} $|$ \texttt{other} \\
\texttt{other\_label} & \textit{string} & opt &
  Required when \texttt{type=`other'}; descriptive label for custom type \\
\end{tabular}

\medskip
\noindent\textbf{\textcolor{schemeaccent}{Target}}\\[2pt]
\noindent\textit{The entity/role subject to the obligation, prohibition, or
permission — WHO must comply, not the recipient or intermediary.}\\[2pt]
\begin{tabular}{@{}l l l p{5.2cm}@{}}
\texttt{role}       & \textit{string} & \textbf{req} & Role/entity subject to the rule \\
\texttt{qualifiers} & \textit{string} & opt           & Narrowing qualifiers (e.g.\ ``foreign private issuer'') \\
\end{tabular}

\medskip
\noindent\textbf{\textcolor{schemeaccent}{Citations}}\\[2pt]
\begin{tabular}{@{}l l l p{5.2cm}@{}}
\texttt{text} & \textit{string $|$ string[ ]} & opt & Textual citation(s) to supporting sources \\
\end{tabular}
\end{tcolorbox}
\vspace{1mm}
{\small\sffamily\textit{Schema~2: Individual rule unit schema with detailed type taxonomy
(11~categories), target identification, and citation support.}}

\bigskip

\begin{tcolorbox}[
  schemebox,
  title={\textbf{\small\sffamily\color{white} Schema~3
    \textnormal{\color{white!80!schemetitle}$\mid$}
    Statement \textnormal{\color{white!70!schemetitle}within a rule unit}}},
]
\bigskip
\noindent Decomposition of a regulatory requirement into action, object, method,
constraints, conditions, exceptions, and verbatim source text.

\medskip
\noindent\textbf{\textcolor{schemeaccent}{Statement}}\\[2pt]
\begin{tabular}{@{}l l l p{4.8cm}@{}}
\texttt{action}           & \textit{string}                        & \textbf{req} & Primary regulatory action as verb phrase \\
\texttt{action\_object}   & \textit{string}                        & opt           & Direct object or recipient of the action \\
\texttt{method}           & \textit{string}                        & opt           & How the action must be performed \\
\texttt{constraints}      & \textit{Constraint[ ]}                 & \textbf{req} & Limits imposed on applying this rule (default \texttt{[]}) \\
\texttt{conditions}       & \textit{Condition[ ]}                  & \textbf{req} & Prerequisites for this rule to apply (default \texttt{[]}) \\
\texttt{exceptions}       & \textit{ExceptionItem[ ]}              & \textbf{req} & Places where this rule may not apply (default \texttt{[]}) \\
\texttt{penalties\_or\_}  & \textit{PenaltiesOr-}                  & opt           & Penalties or consequences \\
\quad\texttt{consequences}& \textit{Consequences[ ]}               &               & of a rule action \\
\texttt{purpose}          & \textit{string}                        & opt           & Stated purpose/objective if explicit in source \\
\texttt{verbatim}         & \textit{string}                        & \textbf{req} & Exact quoted text from the source \\
\end{tabular}

\medskip
\noindent\textbf{\textcolor{schemeaccent}{Constraint}}\\[2pt]
\begin{tabular}{@{}l l l p{5.2cm}@{}}
\texttt{text}        & \textit{string} & \textbf{req} & Constraint description \\
\texttt{applies\_to} & \textit{string} & opt           & Entity the constraint binds to \\
\end{tabular}

\medskip
\noindent\textbf{\textcolor{schemeaccent}{Condition}}\\[2pt]
\begin{tabular}{@{}l l l p{5.2cm}@{}}
\texttt{trigger}           & \textit{string}       & \textbf{req} & Triggering event or conditional text \\
\texttt{time\_window}      & \textit{TimeWindow}   & opt           & Temporal boundaries for the condition \\
\texttt{cross\_references} & \textit{CrossRef[ ]}  & opt           & Related regulatory cross-references \\
\end{tabular}

\medskip
\noindent\textbf{\textcolor{schemeaccent}{TimeWindow}}\\[2pt]
\begin{tabular}{@{}l l l p{5.2cm}@{}}
\texttt{start} & \textit{string} & opt & Start of the time window \\
\texttt{end}   & \textit{string} & opt & End of the time window \\
\texttt{zone}  & \textit{enum}   & opt & \texttt{ET} $|$ \texttt{EST} $|$ \texttt{EDT} \\
\end{tabular}

\medskip
\noindent\textbf{\textcolor{schemeaccent}{ExceptionItem}}\\[2pt]
\begin{tabular}{@{}l l l p{5.2cm}@{}}
\texttt{text}              & \textit{string}      & \textbf{req} & Main exception description \\
\texttt{cross\_references} & \textit{CrossRef[ ]} & opt           & Related regulatory references \\
\end{tabular}

\medskip
\noindent\textbf{\textcolor{schemeaccent}{PenaltiesOrConsequences}}\\[2pt]
\begin{tabular}{@{}l l l p{5.2cm}@{}}
\texttt{text}              & \textit{string}      & \textbf{req} & Penalty/consequence description \\
\texttt{cross\_references} & \textit{CrossRef[ ]} & opt           & Related regulatory references \\
\end{tabular}

\medskip
\noindent\textbf{\textcolor{schemeaccent}{CrossRef}} \\[2pt]
\begin{tabular}{@{}l l l p{5.2cm}@{}}
\texttt{type}     & \textit{enum}   & \textbf{req} & \texttt{CFR} $|$ \texttt{Rule} $|$ \texttt{Form} $|$ \texttt{USC} $|$ \texttt{Release} $|$ \texttt{Regulation} $|$ \texttt{Note} $|$ \texttt{Other} \\
\texttt{cite}     & \textit{string} & \textbf{req} & Citation string \\
\texttt{summary}  & \textit{string} & opt           & Brief summary of the cross-reference \\
\end{tabular}
\end{tcolorbox}
\vspace{1mm}
{\small\sffamily\textit{Schema~3: Statement decomposition with conditions,
constraints, exceptions, penalties, and cross-references.
Arrays default to~\texttt{[]}; optional scalars default to~\texttt{null}.}}

\bigskip

\begin{tcolorbox}[
  schemebox,
  title={\textbf{\small\sffamily\color{white} Schema~4
    \textnormal{\color{white!80!schemetitle}$\mid$}
    Judge Evaluation Schemas \textnormal{\color{white!70!schemetitle}(Judges 1, 2, 3)}}},
]
\bigskip
\noindent Multi-step evaluation framework for extraction quality. Each metric
carries an integer \texttt{Score}~$\in [0,5]$ and a textual
\texttt{Justification}. All fields in every judge step are \textbf{required}.

\medskip
\noindent\textbf{\textcolor{schemeaccent}{Metric}} \textit{(shared unit)}\\[2pt]
\begin{tabular}{@{}l l l p{5.2cm}@{}}
\texttt{Score}         & \textit{integer} & \textbf{req} & $0 \le \texttt{Score} \le 5$ \\
\texttt{Justification} & \textit{string}  & \textbf{req} & Textual justification for the score \\
\end{tabular}

\medskip
\noindent\textbf{\textcolor{schemeaccent}{Step1Judge}}
\textit{--- Section-level metadata extraction}\\[2pt]
\begin{tabular}{@{}l p{7.8cm}@{}}
\texttt{Completeness}                                & Coverage of all required metadata fields \\
\texttt{Fidelity\_to\_source\_text}                  & Faithfulness to the original text \\
\texttt{Non\_hallucination}                          & Absence of fabricated content \\
\texttt{Title\_Quality}                              & Appropriateness and clarity of the extracted title \\
\texttt{Precision\_of\_Citations\_and\_Dates}        & Accuracy of citation strings and date formats \\
\texttt{Reasonable\_Population\_of\_Optional\_Fields} & Sensible use of optional fields \\
\end{tabular}

\medskip
\noindent\textbf{\textcolor{schemeaccent}{Step2Judge}}
\textit{--- Definition extraction}\\[2pt]
\begin{tabular}{@{}l p{7.8cm}@{}}
\texttt{Completeness}                      & All defined terms captured \\
\texttt{Fidelity\_to\_Source\_Text}        & Definitions faithful to source \\
\texttt{No\_Hallucination\_or\_Fabrication} & No invented terms or definitions \\
\texttt{Precision\_and\_Formatting}        & Correct formatting and precision \\
\texttt{Quality\_of\_Terms}                & Appropriateness and clarity of extracted terms \\
\end{tabular}

\newpage
\medskip
\noindent\textbf{\textcolor{schemeaccent}{Step3Judge}}
\textit{--- Per-rule unit evaluation}\\[2pt]
\begin{tabular}{@{}l p{7.8cm}@{}}
\texttt{Completeness}              & All rule components present \\
\texttt{Conciseness}               & Language is brief while preserving meaning \\
\texttt{Accuracy}                  & Rule type correctly classifies the source \\
\texttt{Consistency}               & Targets align with the source \\
\texttt{Fidelity\_to\_source\_text} & Statement reflects the source faithfully \\
\texttt{Neutrality}                & Balanced, unbiased presentation \\
\texttt{Actionability}             & Clear, usable guidance with reasonable abstraction and minimal ambiguity \\
\texttt{Non\_hallucination}        & No fabricated rule components \\
\end{tabular}

\medskip
\noindent\textbf{\textcolor{schemeaccent}{JudgeScore}} \textit{(aggregate)}\\[2pt]
\begin{tabular}{@{}l l l p{5.2cm}@{}}
\texttt{step1} & \textit{Step1Judge} & \textbf{req} & Step~1 evaluation \\
\texttt{step2} & \textit{Step2Judge} & \textbf{req} & Step~2 evaluation \\
\texttt{step3} & \textit{Step3Judge} & \textbf{req} & Step~3 evaluation \\
\texttt{Final} & \textit{integer}    & opt           & Overall (average) score $\in [0,5]$ \\
\texttt{Notes} & \textit{string}     & opt           & Free-text evaluator notes \\
\end{tabular}
\end{tcolorbox}
\vspace{1mm}
{\small\sffamily\textit{Schema~4: Three-step judge framework ---
Step~1 evaluates metadata (6~metrics), Step~2 evaluates definitions (5~metrics),
Step~3 evaluates each rule unit (8~metrics).
All metrics use the shared \texttt{Metric} type with \texttt{Score} $\in [0,5]$
and \texttt{Justification}.}}

\section{Sample Evaluation Questions}
\label{appx:sample_questions}

Table~\ref{tab:sample_questions} presents a representative sample of 10 questions
from the 100-question evaluation set used in Experiment~3
(Section~\ref{sec:downstream-eval}), generated by Qwen3-VL-8B-Instruct
from HIPAA regulatory passages using the prompt in
Appendix~\ref{appx:question_gen_prompt}. The questions span five distinct HIPAA
topic areas (general use and disclosure standards, permitted uses, business associate
obligations, genetic information protections, and reproductive health privacy) and
mix factual, conditional, and analytical question types, collectively stress-testing
whether a RAG system can retrieve and reason over structurally and semantically
diverse regulatory provisions.

\begin{table}[h]
\centering
\renewcommand{\arraystretch}{1.3}
\setlength{\tabcolsep}{6pt}
\begin{tabular}{p{0.04\linewidth} p{0.88\linewidth}}
\toprule
\textbf{\#} & \textbf{Question} \\
\midrule
1  & Who may not use or disclose protected health information except as permitted or
     required by this subpart? \\
2  & When is a covered entity permitted to use or disclose protected health information
     for treatment, payment, or health care operations? \\
3  & What conditions must be met for a use or disclosure to be considered incident to
     a permitted use or disclosure? \\
4  & What are the required disclosures of protected health information by a covered
     entity? \\
5  & What are the limitations on a business associate's use and disclosure of protected
     health information? \\
6  & What is prohibited regarding the use and disclosure of genetic information for
     underwriting purposes? \\
7  & What activities are considered ``underwriting purposes'' with respect to genetic
     information? \\
8  & What constitutes a ``sale of protected health information''? \\
9  & Under what conditions does the prohibition on using reproductive health care
     information apply? \\
10 & What is presumed about the lawfulness of reproductive health care provided by
     another person, and what can override that presumption? \\
\bottomrule
\end{tabular}
\caption{Ten representative questions from the 100-question evaluation set used in
Experiment~3. Each question is fully answerable from its source passage alone.
The set spans five HIPAA topic areas and three question types, reflecting the
diversity requirements specified in the generation prompt
(Appendix~\ref{appx:question_gen_prompt}).}
\label{tab:sample_questions}
\end{table}
\end{document}